\documentclass[default,iicol]{sn-jnl}

\usepackage{subfig}
\usepackage{pifont}
\usepackage{colortbl}
\usepackage[export]{adjustbox}
\usepackage{arydshln}
\usepackage{xspace}
\usepackage[
acronyms,
nohypertypes={acronym},
shortcuts,
nonumberlist,
nogroupskip,
nopostdot
]{glossaries}
\usepackage[sort&compress,capitalize,nameinlink]{cleveref}

\usepackage{graphicx}%
\usepackage{multirow}%
\usepackage{amsmath,amssymb,amsfonts}%
\usepackage{amsthm}%
\usepackage{mathrsfs}%
\usepackage[title]{appendix}%
\usepackage{xcolor}%
\usepackage{textcomp}%
\usepackage{manyfoot}%
\usepackage{booktabs}%
\usepackage{algorithm}%
\usepackage{algorithmicx}%
\usepackage{algpseudocode}%
\usepackage{listings}%
\graphicspath{{./figures/}}

\theoremstyle{thmstyleone}%
\theoremstyle{thmstyletwo}%

\theoremstyle{thmstylethree}%
\newacronym{fps}{FPS}{Farthest-Point-Sampling}
\newacronym{rs}{RS}{Random-Sampling}
\newacronym{ft3d}{$\mathrm{FT3D_s}$}{FlyingThings3D subset}
\newacronym{ft3do}{$\mathrm{FT3D_o}$}{FlyingThings3D}

\newacronym{ft3d_bold}{$\mathrm{\mathbf{FT3D_s}}$}{FlyingThings3D Subset_}

\newacronym{lfa}{LFA}{Local-Feature-Aggregation}
\newacronym{knn}{KNN}{K-Nearest-Neighbor}
\newacronym{us}{US}{Upsampling}
\newacronym{ds}{DS}{Downsampling}
\newacronym{fe}{FE}{Flow-Embedding}
\newacronym{wl}{WL}{Warping-Layer}
\newacronym{max}{max-pooling}{max-pooling}
\newcommand\Tstrut{\rule{0pt}{2.1ex}}         
\newcommand\Bstrut{\rule[-0.9ex]{0pt}{0pt}}   
\newcommand{\cmark}{\ding{51}}%
\newcommand{\xmark}{\ding{55}}%

\definecolor{Gray}{gray}{0.85}
\definecolor{LightCyan}{rgb}{0.88,1,1}

\newcolumntype{a}{>{\columncolor{Gray}}c}
\newcolumntype{b}{>{\columncolor{white}}c}
\newcolumntype{M}[1]{>{\centering\arraybackslash}m{#1}}

\newcommand*{\eg}{\textit{e.g.}\@\xspace}
\newcommand*{\ie}{\textit{i.e.}\@\xspace}
\newcommand*{\etal}{\textit{et al.}\@\xspace}
\newcommand*{\cf}{\textit{cf.}\@\xspace}

\newcommand{\name}{\mbox{RMS-FlowNet++}} 
\newcommand{\PTDP}{\mbox{\textit{Patch-to-Dilated-Patch}}}
\newcommand{\mytilde}{{\raise.17ex\hbox{$\scriptstyle\sim$}}}

\begin{document}

\title[Article Title]{
	\name{}: Efficient and Robust Multi-Scale Scene Flow Estimation for Large-Scale Point Clouds}


\author*[1]{\fnm{Ramy} \sur{Battrawy}}\email{ramy.battrawy@dfki.de}

\author[1]{\fnm{Ren{\'e}} \sur{Schuster}}\email{rene.schuster@dfki.de}

\author[1]{\fnm{Didier} \sur{Stricker}}\email{didier.stricker@dfki.de}

\affil[1]{\orgdiv{Augmented Vision}, \orgname{German Research Center for Artificial Intelligence (DFKI)}, \orgaddress{\street{Trippstadter Str.~122}, \city{Kaiserslautern}, \postcode{67663}, \state{Rhineland-Palatinate}, \country{Germany}}}

\abstract{
	The proposed \name{} is a novel end-to-end learning-based architecture for accurate and efficient scene flow estimation that can operate on high-density point clouds.
	For hierarchical scene flow estimation, existing methods rely on expensive \acrfull{fps} to sample the scenes, must find large correspondence sets across the consecutive frames and/or must search for correspondences at a full input resolution.
	While this can improve the accuracy, it reduces the overall efficiency of these methods and limits their ability to handle large numbers of points due to memory requirements. 
	In contrast to these methods, our architecture is based on an efficient design for hierarchical prediction of multi-scale scene flow.
	To this end, we develop a special flow embedding block that has two advantages over the current methods: First, a smaller correspondence set is used, and second, the use of \acrfull{rs} is possible. 
	In addition, our architecture does not need to search for correspondences at a full input resolution.
	Exhibiting high accuracy, our \name{} provides a faster prediction than state-of-the-art methods, avoids high memory requirements and enables efficient scene flow on dense point clouds of more than 250K points at once.
	Our comprehensive experiments verify the accuracy of \name{} on the established FlyingThings3D data set with different point cloud densities and validate our design choices.
	Furthermore, we demonstrate that our model has a competitive ability to generalize to the real-world scenes of the KITTI data set without fine-tuning.
}





\maketitle

\section{Introduction} \label{introduction} \glsresetall
Robust perception of the dynamic environment is a fundamental task for many real-world applications such as autonomous driving, robot navigation, augmented reality, and human-robot interaction systems.    	
The goal of scene flow is to estimate 3D displacement vectors between two consecutive scenes, representing all observed points in the scene as a dense or semi-dense 3D motion field. 
Therefore, scene flow can serve as an upstream step in high-level challenging computer vision tasks such as object tracking, odometry, action recognition, etc.
With prior knowledge of the camera's intrinsic parameters, the 3D scene flow can be projected onto the image plane to obtain its 2D counterpart in pixel coordinates, which is called optical flow.

Many approaches propose pixel-wise scene flow estimation using stereo image sequences by combining geometry reconstruction with optical flow estimation to obtain dense scene flow \cite{franke20056d, huguet2007variational, wedel2008efficient, ilg2018occlusions, chen2020consistency, schuster2020sceneflowfields++, schuster2017sceneflowfields, saxena2019pwoc}.
Despite significant advances in such approaches, the overall accuracy of the resulting scene flow is highly dependent on the image quality, which can be poor under adverse lighting conditions.
Compared to stereo systems, LiDAR sensors can accurately capture 3D geometry in the form of 3D point clouds and are less sensitive to lighting conditions. Therefore, there is an increasing emphasis on estimating scene flow directly from 3D point clouds.  

Handling point clouds and finding correspondences in 3D space is more challenging due to the irregularity, sparsity of points, and varying point density of the scene. To tackle these challenges, several techniques develop deep neural architectures to estimate scene flow from point clouds.
Some of these methods project the point cloud onto a permutohedral lattice \cite{gu2019hplflownet} and then use bilateral convolutions \cite{jampani2016learning}. 
Others organize 3D point clouds into voxels \cite{gojcic2021weakly, li2021sctn} and use sparse convolutions \cite{choy20194d} to facilitate scene flow prediction.
However, these regular representations can introduce discretization artifacts and information loss that negatively affect the accuracy of the network. 

With the advent of point-based networks on 3D point clouds \cite{wu2019pointconv, qi2017pointnet++}, many works estimate scene flow directly from \textit{raw} point clouds using the multi-layer perceptron (MLP) as in \cite{liu2019flownet3d, wei2020pv, wu2020pointpwc, kittenplon2021flowstep3d, wang2021hierarchical, gu2022rcp, wang2022residual, wang2022matters, cheng2022bi} without the need for regular or intermediate representations.  
All of these techniques build a flow embedding module at coarse resolutions and then either use hierarchical refinement modules along with upsampling~\cite{cheng2022bi, liu2019flownet3d, wu2020pointpwc, wang2021hierarchical, wang2022residual, wang2022matters} or use gated recurrent units (GRUs) \cite{cho2014learning} with iterative flow updating for the refinement process \cite{wei2020pv, kittenplon2021flowstep3d, gu2022rcp}. Despite their ability to capture both near and far matches, GRU-based methods \cite{wei2020pv, kittenplon2021flowstep3d, gu2022rcp} are less efficient in terms of runtime due to iterative flow updates along with expensive flow embedding layers.

\begin{figure}[t]
	\begin{center}
		\includegraphics[width=1.0\linewidth]{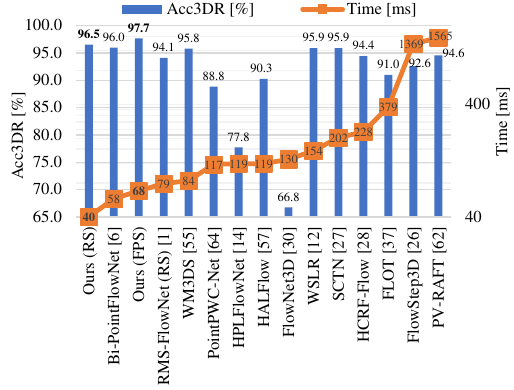}
		\caption{Our \name{} shows an accurate scene flow (Acc3DR) with a low runtime. The accuracy is tested on $\mathrm{KITTI_s}$~\cite{menze2015object} with $8192$ points as input and the runtime is analyzed for all methods equally on a Geforce GTX 1080 Ti.}
		\label{Figure1_Teaser}
	\end{center}
\end{figure}

Following hierarchical schemes, Wang \etal~\cite{wang2022residual} propose a double attentive flow embedding along with the explicit learning of the residual scene flow. 
Their extension in \cite{wang2022matters} further improves the results by jointly learning of backward constraints in the all-to-all flow embedding layer to capture distant matches. 
However, both methods \cite{wang2022matters, wang2022residual} reduce the scene flow resolution to a quarter of the input points, and they show that obtaining high accuracy of the scene flow at the full input resolution requires a further refinement module. This can be computationally expensive, while using a simple interpolation method can degrade the overall accuracy. 
In addition, the use of an all-to-all flow embedding layer in \cite{wang2022matters} increases the size of the correlation volume, which in turn increases the computational load of further operations.
More recently, Bi-PointFlowNet \cite{cheng2022bi} has proposed to learn bidirectional correlations from coarse-to-fine, searching for correspondences in both directions, and it uses a flow embedding layer at full input resolution.

All of the above point-based methods use~\gls{fps} and rely on a~\gls{knn} search with a large set of correspondences during the flow embedding, which both increases the computational and memory requirements and limits the ability to handle large point clouds.

\begin{figure*}[t]
	\begin{center}
		\includegraphics[width=1.0\linewidth]{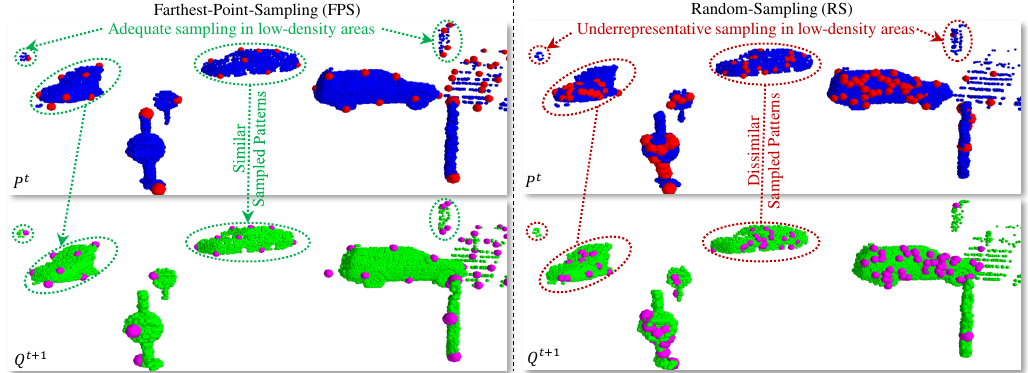}
		\caption{The challenges of \acrfull{rs} (right) compared to \acrfull{fps} (left): Both techniques sample two consecutive scenes $P^t$ (blue) and $Q^{t+1}$ (green) into red and pink samples, respectively. Areas of low density are often not sufficiently covered by \gls{rs}, resulting in dissimilar patterns. The patterns of the corresponding objects are much more similar when \gls{fps} is used, making it easier to match the points.}
		\label{Figure2_FPSvsRS}
	\end{center}
\end{figure*}

To tackle these challenges, we present our \name{} -- a hierarchical point-based learning approach that requires smaller correspondence sets compared to the state-of-the-art methods and outperforms them when using \gls{fps} on the KITTI \cite{menze2015object} data set.
In addition, our model allows the use of \Gls{rs} and is therefore more efficient, has a smaller memory footprint, and shows comparable results at a lower runtime compared to the other state-of-the-art methods (\cf~\cref{Figure1_Teaser}).

The advantage of \gls{rs} combined with the smaller correspondence set results in our model being the only one that can robustly estimate scene flow on a very large set of points, as shown in \cref{Density}.
However, using \gls{rs} for scene flow estimation is challenging for two main reasons:
1.) \Gls{rs} will reflect the spatial distribution of the input point cloud, which is problematic if it is far from uniform, which is a disadvantage compared to \Gls{fps}.
2.) Corresponding (rigid) areas between consecutive point clouds will be sampled differently by \gls{rs}, while \gls{fps} will yield more similar patterns.
Both issues are illustrated in \cref{Figure2_FPSvsRS}.

To overcome these problems, we propose a novel \PTDP{} flow embedding consisting of four layers with lateral connections (\cf~\cref{Figure6_FlowEmbedding}) to incorporate a larger receptive field during matching without increasing the physical set of correspondences.
Overall, our fully supervised architecture consists of a hierarchical feature extraction, an optimized flow embedding, and scene flow prediction at multiple scales.
The preliminary version of our network design has been published in RMS-FlowNet~\cite{battrawy2022rms}, but we are improving the overall design, which will lead to a very accurate result with higher efficiency.
Our contribution can be summarized as follows:
\begin{itemize}
	\item We propose \name{} -- an end-to-end scene flow estimation network that operates on dense point clouds with high accuracy.
	\item Our network consists of a hierarchical scene flow estimation with a novel flow embedding module (called \PTDP{}) which is suitable for the combination with \acrlong{rs}.
	\item Compared to our previous work in RMS-FlowNet \cite{battrawy2022rms}, we significantly reduce the size of the correspondence set, and omit some layers in the feature extraction module to increase the overall efficiency. Furthermore, we show that a feature-based search can increase the overall accuracy without sacrificing the efficiency.
	\item We explore the advantages of \gls{rs} over \gls{fps} on high-density point clouds and its ability to generalize during the inference.
	\item We provide an intensive benchmark showing our strong results in terms of accuracy, generalization, and runtime compared to previous methods. 
	\item Finally, we investigate the robustness of our network to occlusions and evaluate it for points acquired at longer distances ($>$ 35 m).
\end{itemize}

\section{Related Work}

\begin{figure*}[t]
	\begin{center}
		\includegraphics[width=1.0\linewidth]{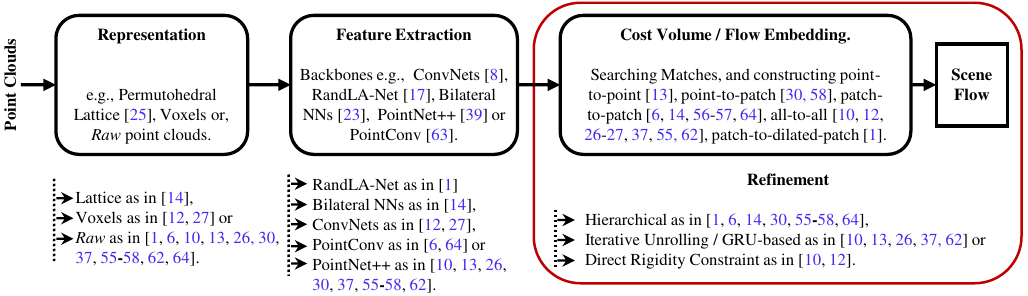}
		\caption{We describe the generic pipeline of recent scene flow estimation methods. Like our previous work \cite{battrawy2022rms}, our \name{} estimates scene flow directly from \textit{raw} point clouds and extracts features based on RandLA-Net~\cite{hu2020randla}. Compared to recent scene flow methods, our novel \PTDP{} allows the use of \gls{rs} along with hierarchical or coarse-to-fine refinement.}
		\label{Figure3_GenericPipeline}
	\end{center}
\end{figure*}

3D scene flow was first introduced by \cite{vedula1999three}, developed using image-based (\eg, RGB-D) setups  \cite{hadfield2011kinecting, hornacek2014sphereflow, quiroga2014dense, jaimez2015primal, jaimez2015motion, sun2015layered}, and then further developed in advanced deep learning networks \cite{shao2018motion, teed2021raft}. 
Since RGB-D sensors can only perceive depth at short distances, there have been many works that estimate scene flow from stereo images by jointly estimating disparity and optical flow \cite{ilg2018occlusions, jiang2019sense, ma2019deep, chen2020consistency}. 
However, two-view geometry has inherent limitations in self-driving cars, such as inaccuracies in disparity estimation in distant regions. 
It can also suffer from poor lighting conditions, such as in dark tunnels.
Our work focuses on learning scene flow directly from point clouds, without relying on RGB images.

\textbf{Scene Flow from Point Clouds:} 
With the recent advent of LiDAR sensors, which provide highly accurate 3D geometry of the environment for autonomous driving and robot navigation, it becomes increasingly important to estimate scene flow directly from point clouds in 3D world space. 
In this context, there is some work \cite{dewan2016rigid, ushani2017learning} that formulates the task of scene flow estimation from point clouds as an energy optimization problem without taking advantage of deep learning.
Advances in deep learning on 3D point clouds \cite{qi2017pointnet, qi2017pointnet++, su2018splatnet, wu2019pointconv} make neural networks more attractive and accurate for 3D scene flow estimation than the traditional methods \cite{ushani2018feature, wang2018deep, liu2019flownet3d, behl2019pointflownet, gu2019hplflownet, wu2020pointpwc, puy20flot, wei2020pv, kittenplon2021flowstep3d, li2021hcrf, wang2021hierarchical, gojcic2021weakly, gu2022rcp, cheng2022bi}.
These recent methods mostly follow a general scene flow estimation pipeline as shown in \cref{Figure3_GenericPipeline}, but differ in how they represent point clouds, extract features, design the cost volume, or apply the refinement strategy.
For example, with the breakthrough architecture of PointNet++ \cite{qi2017pointnet++}, many works estimate scene flow directly from \textit{raw} point clouds in an end-to-end fashion \cite{liu2019flownet3d, puy20flot, wei2020pv, kittenplon2021flowstep3d, wang2021festa, wang2021hierarchical, gu2022rcp, wang2022residual, wang2022matters, dong2022exploiting}. 
Based on PointNet++, FlowNet3D \cite{liu2019flownet3d} is the first work to introduce a novel flow embedding layer. 
However, its accuracy is limited because there is only a single flow embedding layer and the correlation in local regions relies on the nearest spatial neighbor search, which may fail for long-range motion (\ie, distant matches). 
In an attempt to overcome the limitations of FlowNet3D, many approaches introduce hierarchical scene flow estimation, iterative unrolling methods, or work under rigidity assumptions.

\textit{Hierarchical Scene Flow:} 
HPLFlowNet \cite{gu2019hplflownet} introduces multi-scale correlation layers by projecting the points into a permutohedral lattice as in SplatNet \cite{su2018splatnet} and applying bilateral convolutional layers (BCL) \cite{kiefel2014permutohedral, jampani2016learning}. 
Despite of the efficiency of HPLFlowNet \cite{gu2019hplflownet} on high-density point clouds, but the accuracy of the network is prone to unavoidable errors due to the splatting and slicing process. 
PointPWC-Net \cite{wu2020pointpwc} avoids the grid representation in \cite{gu2019hplflownet} and improves the scene flow accuracy on \textit{raw} point clouds based on PointConv \cite{wu2019pointconv} by regressing multi-scale flows from coarse-to-fine. 
Following the hierarchical point-based designs, HALFlow \cite{wang2021hierarchical} uses the point feature learning of PointNet++ \cite{qi2017pointnet++}, but proposes a hierarchical attention learning flow embedding with double attentions leading to better results than PointPWC-Net \cite{wu2020pointpwc}. 
Further improvements are proposed in \cite{wang2022residual, wang2022matters} to develop the flow embedding of \cite{wang2021hierarchical} through explicit prediction of residual flow \cite{wang2022residual} and backward reliability validation \cite{wang2022matters}.
All previous methods take advantage of \gls{fps} for downsampling to provide accurate scene flow estimation, but at the cost of efficiency, especially for dense points. 
Compared to these methods, our network solves the challenge of using \gls{rs} to work with high-density points. 

\textit{Iterative Unrolling for Scene Flow:} 
Besides hierarchical flow embedding schemes, a new trend started in FLOT \cite{puy20flot}, inspired by \cite{li2019algorithm, monga2021algorithm, teed2020raft}, to iteratively refine the scene flow by unrolling a fixed number of iterations to globally optimize an optimal transport map \cite{titouan2019optimal}.
PV-RAFT~\cite{wei2020pv}, FlowStep3D \cite{kittenplon2021flowstep3d}, and RCP \cite{gu2022rcp} extend unrolling techniques from optimization problems to learning-based models by using gated recurrent units (GRUs) \cite{cho2014learning} and capturing both local and global correlations.  
We find that iterative unrolling with a fixed number of iterations and repeated use of flow re-embedding works well at low input resolution, but is inefficient compared to hierarchical designs. 

\textit{Rigidity Assumption for Scene Flow:}
Axiomatic concepts of explicit rigidity assumptions with ego-motion estimation are explored in \cite{gojcic2021weakly, dong2022exploiting}.
A plug-in refinement module is proposed by HCRF-Flow \cite{li2021hcrf}, which uses high-order conditional random fields (CRFs) to refine the scene flow by applying the rigidity condition at the region level.
Our \name{} is free of any rigidity constraint, so it can work with non-rigid bodies, such as pedestrians.

\textbf{Flow Embedding on Point Clouds:} The irregular data structure of point clouds makes it difficult to build cost volumes as with image-based solutions \cite{ilg2017flownet2, sun2018pwc, teed2020raft}.
Therefore, previous works such as \cite{gu2019hplflownet, kittenplon2021flowstep3d, liu2019flownet3d, wu2020pointpwc} design complicated flow embedding layers to aggregate the matching costs from consecutive point clouds.

\textit{Patch-to-Point Correlation:} FlowNet3D \cite{liu2019flownet3d} introduces the flow embedding in a patch-to-point manner, which means that the set of neighboring correspondences in the target point cloud set are grouped into the source one based on the Euclidean space. 
Then, the correlations are learned using multi-layer perceptron (MLP) followed by \acrshort{max} to aggregate the features of the correspondence set.

\textit{Patch-to-Patch Correlation:} To incorporate a large field of correlations leading to better accuracy, HPLFlowNet \cite{gu2019hplflownet} proposes a multi-scale patch-to-patch design that takes advantage of the regular representation using the permutohedral lattice  \cite{kiefel2014permutohedral, jampani2016learning}. 
Apart from regular representations, PointPWC-Net \cite{wu2020pointpwc} uses a patch-to-point flow embedding layer to aggregate the features of the correspondence set in the adjacent frames based on the point-wise continuous convolution in PointConv \cite{wu2019pointconv}. 
A point-to-patch embedding is then applied to aggregate the set of neighbor correspondences in the source itself.
Instead of using the backbone of PointConv \cite{wu2019pointconv}, HALFlow \cite{wang2021hierarchical} uses a two-stage of attention mechanism to softly weight the neighboring correspondence features and allocate more attention to the regions with correct correspondences. 
With two-stage attentions, hierarchical and explicit learning of the residual scene flow is proposed by \cite{wang2022residual} to reduce the inconsistencies between the correlations and to handle fast-moving objects. 
Bi-PointFlowNet \cite{cheng2022bi} uses the patch-to-patch mechanism in a bidirectional manner across all multi-scale layers, which requires intensive computation of forward-backward \glspl{knn} and additional refinement at full input resolution.
Compared to \cite{cheng2022bi}, our network finds reliable correlations under bidirectional constraints using the cosine similarity matrix at the coarse scale, and then operates unidirectionally at the upper scales, requiring a small number of correlations defined by the \gls{knn} search. 
It also makes our solution more efficient without sacrificing accuracy by eliminating the need to refine at full input resolution.

\textit{All-to-All Correlations:} There are several approaches that compute a global cosine similarity based on latent features and then learn soft correlations by iteratively refining an optimal transport problem using the non-parametric Sinkhorn algorithm as in FLOT \cite{puy20flot}. 
WSLR~\cite{gojcic2021weakly} uses Sinkhorn, but refines the scene flow at the object-level based on the rigidity assumption and in combination with ego-motion estimation. 
Apart from object-level refinement, FlowStep3D \cite{kittenplon2021flowstep3d} computes an initial global correlation matrix, and then uses gated recurrent units (GRUs) for local region refinement to iteratively align point clouds. 
Another GRU-based method is proposed by PV-RAFT \cite{wei2020pv}, but its flow embedding design combines point-based and voxel-based features to preserve fine-grained information while encoding large correspondence sets at the same time. 
WM3D~\cite{wang2022matters} designs all-to-all correlations, supported by reliability validation, but used only at low scale resolution, where each point in the source uses all points in the target for correlation, and thus each point in the target can therefore obtain the correlation with all points in the source. 
Then, two-stage attentive flow embedding as in \cite{wang2021hierarchical, wang2022residual} is used to aggregate reliable correspondences among this large set of correspondences.
All previous methods significantly increase the number of correlation candidates, which makes the refinement operations computationally intensive, especially when the input point clouds contain a large number of points. 
For this reason, some approaches, such as \cite{wang2021hierarchical, wang2022residual, wang2022matters}, limit the scene flow estimation to the quarter resolution of the input points to avoid further computation.
Compared to \cite{wang2021hierarchical, wang2022residual, wang2022matters}, we design four stages in our flow embedding, ending with two stages of attention for final correlations refinements in a large receptive field. 
It also allows us to get the full resolution of the scene flow exactly the same as the resolution of the input points.

\begin{figure}[t]
	\begin{center}
		\subfloat[][RMS-FlowNet~\cite{battrawy2022rms}.]{\includegraphics[width=0.58\linewidth]{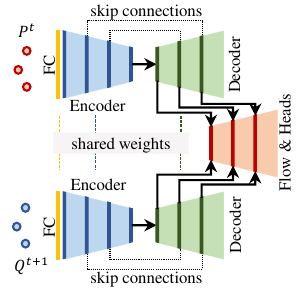}\label{Figure4a_Network_Original}}
		\subfloat[][RMS-FlowNet++.]{\includegraphics[width=0.42\linewidth]{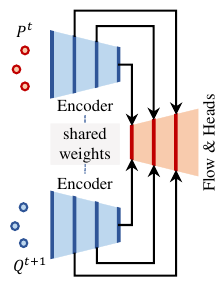}\label{Figure4b_Network_New}}
	\end{center}
	\caption{Our network design consists of feature extraction, flow embedding, warping layers, and scene flow heads, similar to our previous work RMS-FlowNet \cite{battrawy2022rms}. Compared to the feature extraction module in RMS-FlowNet, which consists of fully connected layers (FC) at full input resolution, encoder and decoder modules ($a$), we omit (FC) and the decoder in our \name{} $(b)$.}
	\label{Figure4_Design}
\end{figure}

\textbf{RMS-FlowNet \cite{battrawy2022rms}:} Our preliminary network has proposed a trade-off between efficiency and accuracy by replacing the \gls{fps} sampling technique with the computationally cheap \gls{rs} technique. 
In RMS-FlowNet, we have proposed a novel flow embedding design, called as \PTDP{}, with three embedding steps to solve the challenges of using \gls{rs}.
In addition, we significantly optimize the correspondence search based on \gls{knn} by using the Nanoflann framework \cite{blanco2014nanoflann}\footnote{\url{https://github.com/jlblancoc/nanoflann}}, which further increases the efficiency.
Compared to RMS-FlowNet \cite{battrawy2022rms}, we change the architectural design in our \name{} to speed up the feature extraction module by eliminating the upsampling part (\ie, decoder) (\cf \cref{Figure4_Design}) and the dense layers at the full input resolution in the encoder part. 
And in terms of accuracy, we add another search step in the flow embedding based on the feature space to improve the overall accuracy, and we also improve the way of pairwise correspondence search at the coarse scale under a bidirectional constraint.
With these improvements, our \name{} becomes much more accurate while still showing high efficiency.

\section{Network Design} 
Our \name{} estimates scene flow as translational vectors from consecutive frames of point clouds (\eg, from LiDAR or RGB-D sensors), with no assumptions about object rigidity or direct estimation of sensor motion within the environment (\ie, no direct estimation of ego-motion).

Given Cartesian 3D point cloud frames ${P^{t}=\{p^t_{i}\in\mathbb{R}^3\}}^N_{i=1}$ and ${Q^{t+1}=\{q^{t+1}_{j}\in\mathbb{R}^3\}}^M_{j=1}$ at timestamps $t$ and $t+1$, respectively, our goal is to estimate point-wise 3D flow vectors ${S^t =\{s^t_{i}\in\mathbb{R}^3\}}^N_{i=1}$ for each point within the reference frame $P^{t}$ (\ie, $s^t_{i}$ is the motion vector for $p^t_{i}$).
The sizes ($N$, $M$) of the two frames do not have to be identical, and the two frames should not have exact correspondences between their points.
Our network is designed to estimate scene flow at multi-scale levels through hierarchical feature extraction using a novel design of flow embedding, called \PTDP{}, with warping layers and scene flow estimation heads.

The components of each module are described in detail in the following sections.

\subsection{Feature Extraction Module} \label{feature_extraction}
The feature extraction module consists of two pyramid networks with shared parameters for the hierarchical extraction of two feature sets from $P^t$ and $Q^{t+1}$.
Unlike our previous work in RMS-FlowNet~\cite{battrawy2022rms}, the design of this module includes only the encoder parts, while no decoder and no transposed convolutions are required to upsample the extracted features to the full resolution, as shown in~\cref{Figure4_Design}.

The encoder part computes a hierarchy of features at four scales $\{l_k\}^3_{k=0}$ from fine-to-coarse resolution, where $l_0$ is the full resolution of ($P^{t}$ and $Q^{t+1}$) and the resolutions of the downsampled scales are fixed to ${\{\{l_k\}}^3_{k=1}\mid l_1 = 2048, l_2 = 512, l_3 = 128\}$ during training, but are kept adaptive at higher point densities (\cf \cref{Density}).
Each scale is essentially composed of two layers, where \gls{lfa} is applied to aggregate the features at the $l_{k}$ scale, followed by \gls{ds} to aggregate the features from the $l_{k}$ level to $l_{k+1}$, resulting in a decrease in resolution.
Inspired by RandLA-Net \cite{hu2020randla}, which focuses only on semantic segmentation, we use the feature aggregation layer of \gls{lfa}, which consists of three neural units: 1) local spatial encoding to encode the geometric and relative position features, 2) attentive pooling to aggregate the set of neighbor features, and 3) a dilated residual block. 

To apply \gls{lfa}, we search for the number of nearest neighbors ($K_p$) at all scales using \glspl{knn} search in Euclidean space and aggregate the features with two attentive pooling layers designed as in \cite{hu2020randla}, where the attentive pooling unit is based on the mechanism of self-attention \cite{yang2020robust, zhang2019pcan}.
\Gls{ds} samples the points to the defined resolution in layer $l_{k+1}$ and aggregates the nearest neighbors ($K_p$) from the higher resolution $l_{k}$ by using \acrlong{max}. 
During training and evaluation with \gls{rs}, $K_{p}$ is set to $20$ in all layers and changed to $16$ during evaluation with \gls{fps}.

The feature extraction module outputs two feature sets over all scales $\{F_{k}^{t}\in\mathbb{R}^{c_k}\}^3_{k=0}$ and $\{F_{k}^{t+1}\in \mathbb{R}^{c_k}\}^3_{k=0}$ for $\{P_{k}^{t}\in\mathbb{R}^{3}\}^3_{k=0}$ and $\{Q_{k}^{t+1}\in\mathbb{R}^{3}\}^3_{k=0}$, respectively.
Here, $c_k$ is the feature dimension, which is fixed as ${\{\{c\}}^3_{k=0}\mid c_0 = 32, c_1 = 128, c_2 = 256, c_3 = 512\}$.
The feature extraction module of our \name{} is shown in \cref{Figure4b_Network_New} compared to our preliminary design in \cref{Figure4a_Network_Original}.
The design of \gls{lfa} and \gls{ds} allows the use of \gls{rs} but still requires well-designed flow embedding to ensure robust scene flow.

\subsection{Flow Embedding} \label{Flow_embedding}
A flow embedding module across consecutive frames is the key component for correlating the adjacent frames of point clouds, where finding reliable correlations is extremely important for encoding 3D motion.
In this context, previous state-of-the-art methods combine: 1) grouping of correspondences from $Q^{t+1}$, 2) robust aggregation of correspondence features into $P^{t}$, 3) refinement of flow embedding. 

\begin{figure*}[t]
	\begin{center}
		\includegraphics[width=1.0\linewidth]{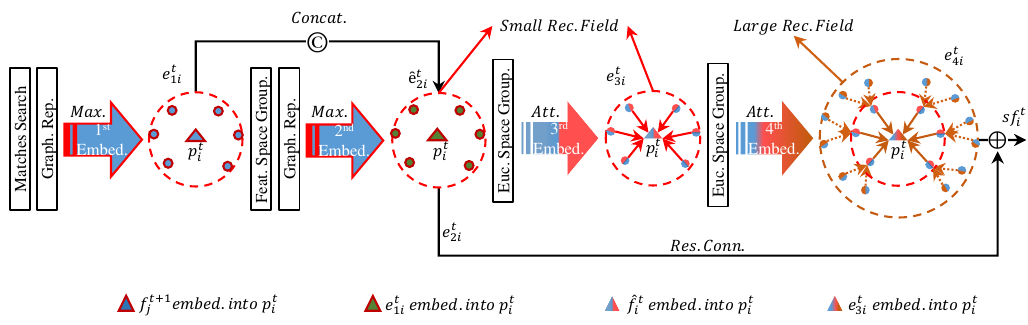}
		\caption{Our novel \acrfull{fe} module consists of four main steps and yields the scene flow feature $sf_i^{t}$: Two maximum embedding layers based on both Euclidean and feature space followed by two attentive embedding layers. Lateral connections are also used: A Concatenation ($Concat.$) between the first two embeddings and a residual connection ($Res.~Conn.$).}
		\label{Figure6_FlowEmbedding}
	\end{center}
\end{figure*}

All point-wise learning-based methods take an advantage of \gls{fps} (as explained in \cref{Figure2_FPSvsRS}) to sample the consecutive frames and rely on finding large correspondence sets ($\geqslant 32$ matches) in the sampled $Q^{t+1}$ based on \gls{knn} as in \cite{wang2021hierarchical, wu2020pointpwc, cheng2022bi} or much more in all-to-all correlations as in \cite{wang2022matters}.
While \gls{fps} can generate similar patterns across consecutive scenes to facilitate obtaining strong match pairs, finding large correspondence sets can increase the likelihood of correlating distant matches (\ie, for large displacements) \cite{wang2022residual, wang2022matters}. 
In addition, some work aims to add refinement at the fine resolution of the input points \cite{wu2020pointpwc, cheng2022bi}.
Taken together, this can increase accuracy, but it reduces the overall efficiency of these methods and limits their ability to handle high point densities. 
In addition, considering large correspondence sets can greatly increase the possibility of aggregating unreliable correlations, leading to inaccurate estimates. 

To address these issues, we develop a special flow embedding module that has two advantages over current point-based methods: First, a smaller correspondence set is used without the need for flow embedding at full resolution, and second, the use of \gls{rs} is possible. 
As a result, we speed up our model and make it amenable to \gls{rs}, as shown by our results in \cref{Experiments}, which allows higher point densities with low memory requirements (\cf \cref{Figure7_AccTimeVsMethods}).
We must recall that, the use of \gls{rs} is more challenging than \gls{fps} (\cf \cref{Figure2_FPSvsRS}) for two reasons.
First, regions with low local point density are underrepresented when using \gls{rs}.
Second, the sampling patterns for corresponding regions are less correlated across frames.

Our novel and efficient flow embedding, called \PTDP{}, aggregates large correspondence sets without increasing the physical number of the nearest neighbors.
This is basically the same design as in our previous work RMS-FlowNet~\cite{battrawy2022rms}, but we apply some changes that are outlined in \cref{Table4_Ablation}.
In this context, we search for correspondences not only in Euclidean space, but also in feature space, and we add another embedding step.

\textbf{Matches Search}: \label{Bidirectional_Map}
Grouping strong correspondences is the first step in any flow embedding. 
Many state-of-the-art methods search for the set of matches based on Euclidean space, but apply soft weights in different ways.  
Since the grouping of correlations based on Euclidean space may not be sufficient to capture distant matches, we use the feature space to find reliable matches at the coarse scale (last down-sampled layer) $l_{3}$.

\textit{Point-to-Point Bidirectional Map:}
For the above reasoning, we compute a simple cosine similarity matrix based on the feature space to find a pair of matches under bidirectional constraint that applies a point-to-point (\ie, one-to-one) correlation map. 
Based on the above reasoning, $q^{t+1}_{j}$ in $Q^{t+1}$ is a true match to $p^t_{i}$ in $P^{t}$ if the highest similarity score is guaranteed in a bidirectional way, otherwise the search for matches is done in Euclidean space.     
Finding robust matches at the coarse scale leads to a high quality initial estimate of the scene flow at scale $l_{3}$. 
This also approximates the distant matches at the upper scales using the warping layer, so that $p^t_{i}$ is close to its match in $Q^{t+1}$.

\textit{Patch-to-Point Search:}
For the upper scales $\{l\}^2_{k=1}$, it is not worth computing the cosine similarity, since it is difficult to get distinctive features in a one-to-one manner at high point densities, and it is worth searching for the number of closest matches $p^t_{i}$ within $Q^{t+1}$ based on the Euclidean space, denoted by $\mathcal{N}_Q(p^t_{i})$. 

\textbf{Graph Representation}: 
After finding the likelihood correspondences, we construct the correlations in a graph form $\mathcal{G}=(\mathcal{V}, \mathcal{E})$, where $\mathcal{V}$ and $\mathcal{E}$ are the vertices and edges, respectively. Then, we apply multi-layer perceptron (MLP):
\begin{equation}
	\begin{aligned}
		h^{\mathcal{E}}_{\Theta}(v_i) = MLP(\{[v_i, v_j-v_i] \mid (i,j) \in \mathcal{E}\})
		\label{Equation1_GraphRepresentation}
	\end{aligned}	
\end{equation}
where $([.,.]$ denotes the concatenation, $v_i$ is a central vertex feature, and $v_j-v_i$ denotes the edge features. 
This representation is compared in \cref{Table4_Ablation} with the original form in our preliminary work RMS-FlowNet \cite{battrawy2022rms}, which omits the $v_{i}$ part in \cref{Equation1_GraphRepresentation} and keeps only the edge features.

\textbf{Flow Embedding Steps}:
Having found the number of possible matches (\ie, $\mathcal{N}_Q(p^t_{i})$) within the representation in the form described above, we apply the flow embedding aggregation steps at each scale $l_k$ for every $\mathrm{i^{th}}$ element within the downsampled scales $\{l\}^3_{k=1}$, except the full-resolution scale $l_{0}$, as follows:
\begin{itemize}
	\item {$1^{st}$ Embedding} (\textit{Patch-to-Point}): We first apply \acrlong{max} to the output of \cref{Equation1_GraphRepresentation} and obtain $e_{1i}^{t}$ as shown in the following equation:
	\begin{equation}
		\begin{aligned}
			e_{1i}^{t} = \underset{{f_{j}^{t+1}\in \mathcal{N}_Q(p_{i}^{t})}}{MAX}(h^{\mathcal{N}_Q(p_{i}^{t})}_{\Theta}(f_i^{t}))
			\label{Equation2_1stEmbedding}
		\end{aligned}	
	\end{equation}
	\item {$2^{nd}$ Embedding} (\textit{Patch-to-Point}): Compared to RMS-FlowNet, we add another embedding step to group correspondences that are semantically similar by applying \gls{knn} in feature space, inspired by the backbone of DGCNN \cite{wang2019dynamic}.
	For this purpose, we group the number of nearest neighbors $\mathcal{N}_E(e_{1i}^{t})$ for each output element of \cref{Equation2_1stEmbedding} based on the feature space and apply the graph form, where $\mathcal{N}_E$ denotes the neighboring features of $e_{1i}^{t}$. Next, we apply \acrlong{max} to obtain $\hat{e}_{2i}^{t}$, which is then channel-wise concatenated with $e_{1i}^{t}$, followed by a multi-layer perceptron (MLP) to obtain $e_{2i}^{t}$: 
	\begin{equation}
		\begin{aligned}
			\hat{e}_{2i}^{t} = \underset{{e_{1j}^{t}\in \mathcal{N}_E(e_{1i}^{t})}}{MAX}(h^{\mathcal{N}_E(e_{1i}^{t})}_{\Theta}(e_{1i}^{t})),
			\label{Equation3_2ndEmbedding}
		\end{aligned}	
	\end{equation}
	\begin{equation}
		\begin{aligned}
			e_{2i}^{t} = MLP([e_{1i}^{t}, \hat{e}_{2i}^{t}])
			\label{Equation4_2ndEmbedding}
		\end{aligned}	
	\end{equation}
	\item {$3^{rd}$ Embedding} (\textit{Point-to-Patch}): Using channel-wise concatenation, we combine the feature $f_i^t$ of $p_i^t$ with the output of \cref{Equation4_2ndEmbedding} on the coarse scale $l_3$, and further combine the upsampled scene flow feature $sf_{i}^{t}$ (computed by \cref{Equation10_residual}) and the upsampled scene flow $s_{i}^{t}$ (computed by the scene flow head in \cref{SceneFlowHead}) on the upper scales as follows: 
	\begin{equation}
		\begin{aligned}
			\hat{f}_{i}^{t} = [f_{i}^{t}, e_{2i}^{t}, sf_{i}^{t}, s_{i}^{t}]
			\label{Equation5_3rdEmbedding}
		\end{aligned}	
	\end{equation}
	Then, we group the nearest features $\hat{f}_{i}^{t}$ based on the Euclidean search $\mathcal{N}_P(p_i^t)$ ($p_i^t$ is the 3D spatial location of $\hat{f}_{i}^{t}$), then compute the attention weights $w_{1i}^{t}$ and sum the weighted features to obtain $e_{3i}^{t}$: 
	\begin{equation}
		\begin{aligned}
			w_{1i}^{t} = g(\hat{f}_{i}^{t}, \textbf{\textit{W}}),
			\label{Equation6_3rdEmbedding}
		\end{aligned}	
	\end{equation}
	\begin{equation}
		\begin{aligned}
			e_{3i}^{t} = \sum_{n=1}^{K_{p}} (\hat{f}_{n}^{t}\cdot w_{1n}^{t})
			\label{Equation7_3rdEmbedding}
		\end{aligned}	
	\end{equation}
	where $g()$ consists of a shared MLP with trainable weights \textbf{\textit{W}} followed by \textit{softmax}. With this attention mechanism, high attention is paid to the well-correlated features, while the less correlated features are suppressed. The attention-based mechanism is generally inspired by \cite{yang2020robust, zhang2019pcan}.
	\item {$4^{th}$ Embedding} (\textit{Point-to-Dilated-Patch}): It repeats the previous step on the output result $e_{3i}^{t}$ with new attention weights $w_{2i}^{t}$ for the nearest features based on Euclidean space. This embedding layer results in an increased receptive field embedding $e_{4i}^{t}$:
	\begin{equation}
		\begin{aligned}
			w_{2i}^{t} = g(e_{3i}^{t}, \textbf{\textit{W}}),
			\label{Equation8_4rdEmbedding}
		\end{aligned}	
	\end{equation}
	\begin{equation}
		\begin{aligned}
			e_{4i}^{t} = \sum_{n=1}^{K_{p}} (e_{3n}^{t}\cdot w_{2n}^{t})
			\label{Equation9_4rdEmbedding}
		\end{aligned}	
	\end{equation}
	where $g()$ consists of a shared MLP with trainable weights \textbf{\textit{W}} followed by \textit{softmax}. Technically, we aggregate features from a larger range by repeating the aggregation mechanism without physically increasing of the number of the nearest neighbors, inspired by \cite{hu2020randla}.
\end{itemize}
Finally, to improve the quality of our flow embedding, we add a residual connection (Res. Conn.), which is an element-wise summation of $e_{2i}^{t}$ and $e_{4i}^{t}$, resulting in the scene flow feature $sf_{i}^{t}$ (\cf \cref{Figure6_FlowEmbedding}):
\begin{equation}
	\begin{aligned}
		sf_{i}^{t} = e_{2i}^{t} + e_{4i}^{t}
		\label{Equation10_residual}
	\end{aligned}	
\end{equation}
Note that for all of the above flow embedding steps, we need to group a certain number of features ($K_p$) and aggregate their features either by \acrlong{max} or by attention. 
$K_p$ is set to $20$ in all layers with \gls{rs} and changed to $16$ during the evaluation with \gls{fps}.
We found that training with \gls{rs} generalizes well with \gls{fps} without any fine-tuning (\cf \cref{Table6_Ablation_FPSvsRS}).
In addition, we must emphasize that the third and forth embedding steps do not require a new \gls{knn} search because we reuse the predefined neighbors of the feature extraction module.
Together, these four steps lead to our novel \PTDP{} embedding, which is described in \cref{Figure6_FlowEmbedding}.
In this way, we are able to obtain a larger receptive field with a small number of nearest neighbors, which is computationally more efficient.

Experiments with the $1^{st}$, $3^{rd}$, and $4^{th}$ flow embedding steps were performed in our previous work \cite{battrawy2022rms}, and we explore our feature-based search in the additional $2^{nd}$ embedding step in \cref{Table4_Ablation}.

\subsection{Multi-Scale Scene Flow Estimation} \label{SceneFlowHead}
\begin{figure}[t]
	\begin{center}
		\includegraphics[width=1.0\linewidth]{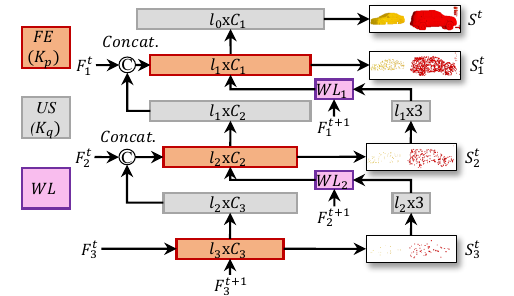}
		\caption{Multi-scale scene flow prediction with three \acrfull{fe} modules (each consists of four steps), two \acrfullpl{wl}, four scene flow estimators and \acrfull{us} layers.}
		\label{Figure5_MultiScale_Prediction}
	\end{center}
\end{figure}
Our \name{} predicts scene flow at multiple scales, inspired by PointPWC-Net \cite{wu2020pointpwc}, but we consider significant changes in conjunction with \gls{rs} to make our prediction more efficient.
Our scene flow prediction over all scales consists of two \glspl{wl}, three \glspl{fe}, four scene flow estimators, and \acrfull{us} modules, as shown in \cref{Figure5_MultiScale_Prediction}.
Compared to the design of PointPWC-Net \cite{wu2020pointpwc} and Bi-PointFlowNet~\cite{cheng2022bi}, we save one element from each category and do not use \gls{fe} at full input resolution. 
As a result, we speed up our model without sacrificing accuracy, as shown in \cref{Figure1_Teaser}.
The multi-scale estimation starts at the coarse resolution by predicting $S_3^t$ with a scene flow estimation module after an initial \gls{fe}.
The scene flow estimation head takes the resulting scene flow features in \cref{Equation10_residual} and applies three layers of MLPs with 64, 32, and 3 output channels, respectively.
Then, the estimated scene flow and the upcoming features from \gls{fe} are upsampled to the next higher scale using one nearest neighbor based on \acrshort{knn} search (\ie, $K_q=1)$. 
We use the same strategy to upsample the scene flow from the $l_1$ scale to the full input resolution $l_0$ without any additional \gls{fe}.

Our \acrlong{wl} uses the upsampled scene flow $S_k^t$ at scale level $l_k$ to warp $P_k^{t}$ toward $Q_k^{t+1}$ to obtain ${\widetilde{P}}_k^{t+1}$. 
This forward warping process does not require any further \gls{knn} search because the predicted scene flow is associated with $P_k^{t}$.
This is more efficient compared to PointPWC-Net~\cite{wu2020pointpwc} or Bi-PointFlowNet~\cite{cheng2022bi}, which must first associate the scene flow with $Q_k^{t+1}$ using \gls{knn} search in order to warp $Q_k^{t+1}$ to $P_k^{t}$ in the backward direction.

\subsection{Loss Function}
The model is fully supervised at multiple scales, similar to PointPWC-Net \cite{wu2020pointpwc}.
If $S_k^t$ is the predicted scene flow and the ground truth is $S_{GT,k}^t$ at scale $l_k$, then the loss can be written as follows:
\begin{equation}
	\begin{aligned}
		\mathcal{L}(\theta) = \sum_{k=0}^{3} {\alpha}_k \sum_{i=1}^{l_k} {\| s_{ki}^t - s_{GT,ki}^t \|}_2,
		\label{Equation1_TrainingLoss}
	\end{aligned}	
\end{equation}
where ${\|.\|}_2$ denotes the $L_2$-norm and the weights per level are $\{\{{\alpha}_k\}^3_{k=0} \mid {\alpha}_0 = 0.02, {\alpha}_1 = 0.04, {\alpha}_2 = 0.08, {\alpha}_3 = 0.16\}$.

\section{Experiments} \label{Experiments}
We conduct several experiments to evaluate the results of our \name{} for scene flow estimation. 
First, we demonstrate its accuracy and efficiency compared to the state-of-the-art methods. 
Second, we verify our design choice with several analyses.

\subsection{Evaluation Metrics} \label{Data_Metrics}
For a fair comparison, we use the same evaluation metrics as in ~\cite{liu2019flownet3d, wu2020pointpwc, wang2021hierarchical, gu2019hplflownet, puy20flot, kittenplon2021flowstep3d, wei2020pv, gu2022rcp, wang2022residual, cheng2022bi, wang2022matters}. 
Let $S^t$ denotes the predicted scene flow, and $S_{GT}^t$ denotes the ground truth scene flow. 
The evaluation metrics are averaged over all points and computed as follows:
\begin{itemize}
	\item {\textit{EPE3D [m]}}: The 3D end-point error computed in meters as ${\|S^t-S_{GT}^t\|}_2$. 
	\item {\textit{Acc3DS [\%]}}: The strict 3D accuracy which is the ratio of points whose EPE3D $< 0.05~m$ \textbf{or} relative error $< 5\%$. 
	\item {\textit{Acc3DR [\%]}}: The relaxed 3D accuracy which is the ratio of points whose EPE3D $< 0.1~m$ \textbf{or} relative error $< 10\%$. 
\end{itemize}
If a metric is subscripted with $"\mathrm{_{noc}}"$, only the non-occluded points are evaluated, otherwise all input points are considered.

\subsection{Data Sets and Preprocessing} \label{Data_Sets} 
As with state-of-the-art methods, we use the original version $\mathrm{FT3D_o}$ and the subset version $\mathrm{FT3D_s}$ of the established large-scale synthetic data set FlyingThings3D (FT3D)~\cite{mayer2016large}. 
The subset version $\mathrm{FT3D_s}$ differs from the original $\mathrm{FT3D_o}$ by excluding some frames from the original and adding more labels.
This data set provides the ground truth for scene flow represented as disparity changes over consecutive frames, disparity maps for consecutive frames and optical flow components, so that the 3D translation vector of the ground truth for scene flow can be computed.
In addition, the $\mathrm{FT3D_s}$ subset provides occlusion maps on the basis of disparity, future and past motions.

In contrast to FT3D, the KITTI data set~\cite{menze2015object} is a small data set with optical flow labels that consists of real outdoor scenes for autonomous driving applications and provides sparse disparity maps generated by a LiDAR sensor. 
The given second disparity map at timestamp $t+1$ has been aligned with the first frame at timestamp $t$, allowing the computation of 3D translation vectors for scene flow.

\textbf{Point Clouds Generation:} Since the existing scenes of data sets and labels do not provide a direct representation of point clouds (\ie, 3D Cartesian locations), the state-of-the-art methods \cite{liu2019flownet3d, wu2020pointpwc, wang2021hierarchical, gu2019hplflownet, puy20flot, kittenplon2021flowstep3d, wei2020pv, gu2022rcp, wang2022residual, cheng2022bi, wang2022matters} basically generate 3D point cloud scenes for their models using the given calibration parameters in the data sets. 
The generated point clouds are randomly subsampled to be evaluated at a certain resolution (\eg, 8192 points) and shuffled to dissolve correlations between consecutive point clouds.
For this, we use the preprocessing strategies of the pioneering work in FlowNet3D~\cite{liu2019flownet3d} and HPLFlowNet~\cite{gu2019hplflownet}, the latter of which yields non-occluded and exact correlations between the scenes.
We also use other preprocessing strategies to ensure that there are no exact correlations between consecutive point clouds. 
To do this, we use the given consecutive disparity maps in $\mathrm{FT3D_s}$, and for the KITTI data set, we use the de-warped disparity maps of the second frame at timestamp $t+1$ generated by \cite{battrawy2019lidar, rishav2020deeplidarflow}.

All of the above preprocessing mechanisms differ in how the second point cloud $Q^{t+1}$ is generated, which either results in exact correlations or not, and whether occluded points are considered or not. This is summarized in \cref{Table1_PreprocessingDataSets} and described in more detail below.

\textbf{Preprocessing in HPLFlowNet~\cite{gu2019hplflownet}\footnote{\url{https://github.com/laoreja/HPLFlowNet}.}:}
This preprocessing considers the complete set of \gls{ft3d}, which consists of $19640$ labeled scenes available in the training set and all $3824$ frames available in the test split for evaluation. 
Unlike the FlowNet3D preprocessing~\cite{liu2019flownet3d}, this preprocessing removes all the occluded points using occlusion maps provided in \gls{ft3d} and the second point cloud frame $Q^{t+1}$ is generated from disparity change and optical flow labels.
For the KITTI data set, the $142$ labeled scenes of the training split available in the~\textit{raw} KITTI data are preprocessed.
The second frame of the point cloud $Q^{t+1}$ is generated from the second disparity map by warping in 3D space, but without occlusion handling.
The data generated from KITTI by this preprocessing is referred to as $\mathrm{KITTI_s}$. 
The preprocessing in HPLFlowNet results in an exact correlation between the consecutive point clouds and occluded points are not taken into account.

\textbf{Preprocessing in FlowNet3D~\cite{liu2019flownet3d}\footnote{\url{https://github.com/xingyul/flownet3d}.}:} 
Here, the original version of \gls{ft3do} is used, with 20,000 images from the training set and 2,000 images from the test set randomly selected for training and evaluation, respectively.
During preprocessing, many occluded points are included in the data and an occlusion mask is computed for $P^{t}$, since there are no predefined occlusion maps in \gls{ft3do}.
The frames of the point clouds $P^{t}$ and $Q^{t+1}$ are generated directly from the consecutive disparity maps and there are no exact correlations between the consecutive scenes.
For the KITTI data set, this preprocessing considers 150 frames with occlusions, but does not compute an occlusion mask. 
The second frame of the point cloud $Q^{t+1}$ is generated using the second disparity map by a warping process in 3D space with occlusion handling. 
The data generated from KITTI by this preprocessing is referred to as $\mathrm{KITTI_o}$. 

\textbf{Preprocessing with Occlusion Masks:}
In contrast to the preprocessing in HPLFlowNet~\cite{gu2019hplflownet}, we generate both point clouds ($P^{t}$ and $Q^{t+1}$) directly from the consecutive disparity maps of the \gls{ft3d}, resulting in very low correlations and existing occlusions in the scenes.
By using the occlusion maps for disparity change and optical flow of consecutive scenes in forward and backward direction provided by \gls{ft3d}, we omit most of the occlusions in consecutive frames, leaving very few occluded points in all frames. 
These remaining occlusions are due to imperfections in the occlusion masks, and are referred to as \textit{partial occlusions}.
We also generate the same data without filtering any of the occlusions. This version is referred to as \textit{large occlusions}.
The preprocessed data from \gls{ft3d} with partial or large occlusions are referred to as $\mathrm{FT3D_{so}}$.

To generate decorrelated points in KITTI, where the given disparity maps of $t+1$ are aligned to the reference view at timestamp $t$, we use the preprocessing mechanism proposed in \cite{battrawy2019lidar, rishav2020deeplidarflow}. 
In this preprocessing, the ground truth of the optical flow is used to generate $Q^{t+1}$ through a pixel-by-pixel de-warping process for each disparity map of $t+1$ aligned with the reference view, which largely dissolves the correlations between the point cloud scenes.
We consider the $142$ labeled scenes of the training split available in the~\textit{raw} KITTI data for preprocessing.
The de-warped disparity maps can be downloaded directly from the source code of DeepLiDARFlow~\cite{rishav2020deeplidarflow}\footnote{\url{https://github.com/dfki-av/DeepLiDARFlow}.}, and are used to compute the point clouds for both frames (\ie, $P^{t}$ and $Q^{t+1}$).
Given the occlusion maps in KITTI, we can either omit or include occluded points to create \textit{partial} or \textit{large occlusions}.
The data generated from KITTI by this preprocessing is referred to as $\mathrm{KITTI_d}$. 

\begin{table}[t]
	\caption{The preprocessing mechanisms of the data sets differ in how the second point cloud $Q^{t+1}$ is generated and whether occluded points are considered or not.}
	\label{Table1_PreprocessingDataSets}
	\begin{center}
		\resizebox{1.0\linewidth}{!}
		{
			\begin{tabular}{M{1.4cm}|M{1.7cm}|M{0.8cm}M{0.8cm}M{0.8cm}M{0.8cm}}
				\textbf{Data Set} & \textbf{Processing} & \textbf{Corre-lated} & \textbf{None Occ.} & \textbf{Partial Occ.} & \textbf{Large Occ.}
				\Tstrut\Bstrut\\
				\hline
				\multirow{3}{*}{\textbf{FT3D~\cite{mayer2016large}}}
				& $\mathbf{FT3D_{s}}$~\textbf{\cite{gu2019hplflownet}} & \cmark & \cmark & \xmark  & \xmark 
				\Tstrut\Bstrut\\
				& $\mathbf{FT3D_{o}}$~\textbf{\cite{liu2019flownet3d}} & \xmark & \xmark & \xmark  & \cmark 	
				\Tstrut\Bstrut\\
				& $\mathbf{FT3D_{so}}$ \textbf{(ours)} & \xmark & \xmark & \cmark  & \cmark 
				\Tstrut\Bstrut\\
				\hline
				\multirow{3}{*}{\textbf{KITTI~\cite{menze2015object}}} 
				& $\mathbf{KITTI_{s}}$~\textbf{\cite{gu2019hplflownet}} & \cmark & \cmark & \xmark  & \xmark 
				\Tstrut\Bstrut\\
				& $\mathbf{KITTI_{o}}$~\textbf{\cite{liu2019flownet3d}} & \xmark & \xmark & \cmark  & \xmark 
				\Tstrut\Bstrut\\
				& $\mathbf{KITTI_{d}}$~\textbf{\cite{rishav2020deeplidarflow}} & \xmark & \xmark & \cmark  & \cmark
				\Tstrut\Bstrut\\
			\end{tabular}
		}
	\end{center}
\end{table}

\begin{table*}[t]
	\caption{When training with non-occluded data on $\mathrm{FT3D_s}$, we evaluate $8192$ points in non-occluded scenes from $\mathrm{FT3D_s}$ and $\mathrm{KITTI_s}$, and the results are written in white cells.
		The light and dark gray cells contain the results with partial and large occlusions from $\mathrm{FT3D_{so}}$ and $\mathrm{KITTI_d}$. In all cases, we evaluate all input points, \ie $8192$. The best result per data set and column is highlighted in bold and the second best is underlined. Each method marked with (\dag{}) is designed to estimate the scene flow at a quarter resolution of the input points (\ie, $2048$ points out of $8192$).}
	\label{Table2_Comparison}
	
	\begin{center}
		\resizebox{1.\linewidth}{!}
		{
			\begin{tabular}{c|l|ccc|>{\columncolor[gray]{.9}[0pt][38mm]}ccc|>{\columncolor[gray]{.7}[0pt][38mm]}ccc} 
				\textbf{Data}   
				& \multirow{3}{*}{\textbf{Model}} 
				& \multicolumn{3}{c|}{\textbf{without occlusions}}
				& \multicolumn{3}{c|}{\textbf{with partial occlusions}}
				& \multicolumn{3}{c}{\textbf{with large occlusions}}
				\\
				\textbf{Set}   
				&
				& \textbf{EPE3D}  & \textbf{Acc3DS}  & \textbf{Acc3DR}			
				& \textbf{EPE3D}  & \textbf{Acc3DS}  & \textbf{Acc3DR}   			
				& \textbf{EPE3D}  & \textbf{Acc3DS}  & \textbf{Acc3DR}
				\\	
				&   
				& [m] $\downarrow$           & [\%] $\uparrow$           & [\%] $\uparrow$                                           
				& [m] $\downarrow$           & [\%] $\uparrow$           & [\%] $\uparrow$                   
				& [m] $\downarrow$           & [\%] $\uparrow$           & [\%] $\uparrow$       
				\Tstrut\Bstrut\\
				\hline
				\multirow{18}{*}{\rotatebox[origin=c]{90}{$\mathbf{FT3D_{s}} / \mathbf{FT3D_{so}}$~\textbf{\cite{mayer2016large}}}}
				& FlowNet3D~\cite{liu2019flownet3d}   
				& 0.114           & 41.25            & 77.06         
				& 0.197           & 21.46            & 43.17         
				& 0.293           & 8.330            & 31.35
				\Tstrut\Bstrut\\
				& HPLFlowNet~\cite{gu2019hplflownet}   		
				& 0.080       	  & 61.60            & 85.57  
				& 0.131           & 44.22            & 72.33         
				& 0.344       	  & 19.02            & 43.27
				\Tstrut\Bstrut\\
				& PointPWC-Net~\cite{wu2020pointpwc}  
				& 0.059        	  & 73.79            & 92.76		  
				& 0.113           & 65.30            & 83.58           
				& 0.292           & 35.72            & 55.94
				\Tstrut\Bstrut\\
				& FLOT~\cite{puy20flot}  
				& 0.052           & 73.20            & 92.70              		  
				& 0.130           & 37.40            & 69.91           
				& 0.227           & 30.26            & 59.21
				\Tstrut\Bstrut\\
				& WSLR~\cite{gojcic2021weakly}	   
				& 0.052 	   	  & 74.60	         & 93.60           
				& 0.140	          & 55.60 	         & 79.10	          
				& 0.461	          & 28.00            & 49.70
				\Tstrut\Bstrut\\
				& HALFlow\dag~\cite{wang2021hierarchical}  
				& 0.049   	      & 78.50	         & 94.68		  
				& 0.125   	      & 46.11	         & 74.96           
				& 0.313           & 22.67            & 47.41
				\Tstrut\Bstrut\\
				& HCRF-Flow~\cite{li2021hcrf}
				& 0.049  	      & 83.37	         & 95.07				  
				& -	              & -	             & -	          
				& -               & -                & -
				\Tstrut\Bstrut\\  
				& PV-RAFT~\cite{wei2020pv}	 
				& 0.046		      & 81.69            & 95.74                 
				& 0.131	          & 61.19            & 81.27	          
				& 0.441           & 29.85            & 50.53
				\Tstrut\Bstrut\\
				& FlowStep3D~\cite{kittenplon2021flowstep3d}  
				& 0.046       	  & 81.62            & 96.14
				& 0.111       	  & 62.67            & 82.75           
				& 0.361       	  & 30.73            & 51.62
				\Tstrut\Bstrut\\
				& RCP~\cite{gu2022rcp}
				& 0.040	     	  & 85.67	   	     & 96.35                      
				& - 	     	  & -	   	         & -	    	  
				& - 	     	  & -	   	         & -	 
				\Tstrut\Bstrut\\ 
				& SCTN~\cite{li2021sctn}
				& 0.038	      	 & 84.70	   	     & 96.80                      
				& - 	      	 & -	   	         & -	    	  
				& - 	      	 & -	   	         & -	 
				\Tstrut\Bstrut\\ 
				& ResidualFlow\dag~\cite{wang2022residual} 
				& 0.031 		 & 91.39 			 & 97.68  
				& - 	         & -	   	         & -	    	  
				& - 	         & -	   	         & -	
				\Tstrut\Bstrut\\ 
				& WM3D\dag~\cite{wang2022matters} 
				& \textbf{0.028} & \textbf{92.90}    & \textbf{98.17}	            
				& \textbf{0.078} & \textbf{77.66}    & \textbf{89.58}	    	  
				& \textbf{0.267} & \underline{46.49}	 & \textbf{65.81}
				\Tstrut\Bstrut\\
				& Bi-PointFlowNet~\cite{cheng2022bi} 
				& \textbf{0.028} & 91.80        	 & 97.80  
				& 0.085 	     & 75.55	   	     & 87.68	    	  
				& 0.284 	     & 46.04	   	     & 62.04
				\Tstrut\Bstrut\\ 
				& RMS-FlowNet (RS) \cite{battrawy2022rms}	
				& 0.051          & 80.00             & 95.60  
				& 0.104	         & 64.20	         & 84.20      
				& 0.305 	     & 35.50             & 57.20
				\Tstrut\Bstrut\\  
				& RMS-FlowNet (FPS) \cite{battrawy2022rms}	
				& 0.052         & 79.10        		 & 95.60 	  
				& 0.111	     	& 61.30	       		 & 83.00          
				& 0.336 	    & 33.10	       		 & 55.20
				\Tstrut\Bstrut\\
				& \textbf{\name{} (RS)}			
				& 0.033	 	    & 91.00        		 & 97.51	          
				& 0.087  		& 76.24        		 & 88.41
				& 0.301 		& 45.12        		 & 63.43
				\Tstrut\Bstrut\\
				& \textbf{\name{} (FPS)}			
				& \underline{0.029}   		  & \underline{92.41}	         & \underline{98.10}		          
				& \underline{0.081}           & \underline{77.54}	         & \underline{89.17}  
				& \underline{0.285}           & \textbf{47.04}            & \underline{64.94}
				\Tstrut\Bstrut\\
				\hdashline
				\multirow{18}{*}{\rotatebox[origin=c]{90}{\textbf{$\mathbf{KITTI_{s}} / \mathbf{KITTI_{d}}$~\cite{menze2015object}}}}
				& FlowNet3D~\cite{liu2019flownet3d}   
				& 0.177        & 37.38    	 & 66.77        
				& 0.206	 	   & 15.16	     & 44.05          
				& 0.330		   & 13.17	     & 36.32
				\Tstrut\Bstrut\\
				& HPLFlowNet~\cite{gu2019hplflownet}   
				& 0.117        & 47.83       & 77.76       
				& 0.161	       & 40.62	     & 68.37
				& 0.338		   & 31.93       & 52.95
				\Tstrut\Bstrut\\
				& PointPWC-Net~\cite{wu2020pointpwc} 	
				& 0.069        & 72.81    	 & 88.84 	 
				& 0.096        & 74.36	     & 85.91
				& 0.276		   & 59.82	     & 69.81
				\Tstrut\Bstrut\\
				& FLOT~\cite{puy20flot} 
				& 0.055     	& 75.93    	 & 91.00             		  
				& 0.135  	    & 57.54	     & 76.08   
				& 0.318	        & 46.71	     & 63.69
				\Tstrut\Bstrut\\
				& WSLR~\cite{gojcic2021weakly}	
				& 0.042	        & 84.90	     & 95.90				  
				& 0.107 		& 72.70	     & 84.50     
				& 0.344 		& 60.90	     & 70.90
				\Tstrut\Bstrut\\
				& HALFlow~\cite{wang2021hierarchical} 	
				& 0.062  	    & 76.49	   	  & 90.26      
				& 0.133	        & 56.88	      & 77.26
				& 0.250 	    & 52.25	   	  & 67.10
				\Tstrut\Bstrut\\
				& HCRF-Flow~\cite{li2021hcrf}
				& 0.053 	    & 86.31	       & 94.44	 				  
				& - 	        & -	   	       & -	    	  
				& - 	        & -	    	   & -	
				\Tstrut\Bstrut\\   
				& PV-RAFT~\cite{wei2020pv}		
				& 0.051 	 	& 83.24	       & 94.55				  
				& 0.095 	    & 73.63	       & 87.60
				& 0.213 	    & 59.67	       & 73.73
				\Tstrut\Bstrut\\
				& FlowStep3D~\cite{kittenplon2021flowstep3d}  
				& 0.056   	    & 79.94	   	   & 92.58
				& 0.103	        & 69.84	       & 84.48
				& 0.296	        & 58.11	       & 70.31
				\Tstrut\Bstrut\\
				& RCP~\cite{gu2022rcp}
				& 0.048	        & 84.91	   	   & 94.48                      
				& - 	        & -	   	       & -	    	  
				& - 	        & -	   	       & -	     
				\Tstrut\Bstrut\\
				& SCTN~\cite{li2021sctn}    
				& 0.037	 		& 87.30	   	   & 95.90	            
				& - 	        & -	   	       & -	    	  
				& - 	        & -	    	   & -
				\Tstrut\Bstrut\\ 	
				& ResidualFlow\dag~\cite{wang2022residual}
				& 0.035  	    & 89.32 	   & 96.20    
				& - 	        & -	   	       & -	    	  
				& - 	        & -	    	   & -	    
				\Tstrut\Bstrut\\ 
				& WM3D\dag~\cite{wang2022matters} 
				& 0.031 	 	& 90.47	       & 95.80           
				& \underline{0.048} 	    & 83.92	       & \underline{92.34}	    	  
				& \underline{0.215} 	    & 65.09	       & 73.44	
				\Tstrut\Bstrut\\ 
				& Bi-PointFlowNet~\cite{cheng2022bi} 
				& \underline{0.030}         & \underline{92.00}        & 96.00  
				& 0.059 	    & \underline{86.08}	   	   & 91.45    	  
				& 0.233 	    & \underline{68.68}	   	   & 74.52								
				\Tstrut\Bstrut\\ 
				& RMS-FlowNet (RS) \cite{battrawy2022rms}	
				&  0.052        & 83.30       & 94.10	  
				&  0.094	    & 69.50	      & 85.40         
				&  0.259	    & 58.40	      & 71.50
				\Tstrut\Bstrut\\  
				& RMS-FlowNet (FPS) \cite{battrawy2022rms}
				& 0.047        & 88.20         & 95.80		 
				& 0.093	       & 75.60	       & 86.90
				& 0.256	       & 63.90	       & 73.20
				\Tstrut\Bstrut\\
				& \textbf{\name{} (RS)}			
				& 0.035       & 90.23          & \underline{96.53} 		          
				& 0.062       & 83.35          & 92.03
				& 0.236       & 66.95          & \underline{75.13}
				\Tstrut\Bstrut\\
				& \textbf{\name{} (FPS)}				
				& \textbf{0.027}  & \textbf{93.98}	& \textbf{97.67}	         
				& \textbf{0.046}  & \textbf{88.94}	& \textbf{94.50}  
				& \textbf{0.214}  & \textbf{71.42}	& \textbf{77.72}
				\Tstrut\Bstrut\\
			\end{tabular}
		}
	\end{center}
\end{table*}

\subsection{Implementation, Training and Augmentation} \label{Implementation}
As in related work, we train our model twice; once with non-occluded data from $\mathrm{FT3D_s}$, considering all frames in the train split of \acrshort{ft3d} \cite{mayer2016large}, and a second time with $\mathrm{FT3D_o}$, containing 20,000 frames with largely occluded points.
During training, the preprocessed data is randomly subsampled to $8192$ points, where the order of the points is random and the correlation between consecutive frames is dissolved by random selection.
Following related work, we remove points with depths greater than $35$ meters, which retains the majority of moving objects contained.

We use the Adam optimizer with default parameters and train the final version of our model with \gls{rs} for $1260$ epochs.
The final model generalizes well with both \gls{rs} and \gls{fps} sampling methods, and no further training with \gls{fps} is required (\cf \cref{Table6_Ablation_FPSvsRS}).
However, to speed up some experiments, we also train with \gls{fps} for $420$ epochs, which converges faster than the training with \gls{rs}. 
When we report the results of the model trained with \gls{fps}, we highlight ($^*$) next to \gls{fps} to distinguish it from the final model trained with \gls{rs}.

We apply an exponentially decaying learning rate that is initialized at $0.001$ and then decreases at a decaying rate of $0.8$ every $20$ epochs when training with \gls{fps} and every $60$ epochs when training with \gls{rs}. 

We add two types of augmentation: First, we add geometric augmentation, \ie, points are randomly rotated by a small angle around the X, Y, and Z axes, and a random translational offset is added to increase the ability of our model to generalize to KITTI~\cite{menze2015object} without fine-tuning.
Second, when training with non-occluded data from \acrshort{ft3d}, each high-resolution frame is randomly sampled to $8192$ points in each epoch differently. 
However, we do not consider this type of augmentation with the \gls{ft3do} because the processed data set using the established preprocessing strategy in FlowNet3D~\cite{liu2019flownet3d} contains only $8192$ points.
The augmentation increases the ability of our model to generalize to the KITTI data set without fine-tuning (\cf \cref{Table7_Ablation_Augmentation}).

\begin{table}[t]
	\caption{Occluded points are taken into account during training and inference. $\mathrm{FT3D_o}$ and $\mathrm{KITTI_o}$ are generated using FlowNet3D~\cite{liu2019flownet3d} preprocessing.}
	\label{Table3_Occlusions}
	\begin{center}
		\resizebox{1.\linewidth}{!}
		{
			\begin{tabular}{c|l|cccc}
				\textbf{Data}  
				& \multirow{2}{*}{\textbf{Model}} 
				& \textbf{EPE3D}  & $\mathbf{EPE3D_{noc}}$ & $\mathbf{Acc3DS_{noc}}$  & $\mathbf{Acc3DR_{noc}}$
				\\ 
				\textbf{Set}	
				&   
				& [m] $\downarrow$        & [m] $\downarrow$   & [\%] $\uparrow$      & [\%] $\uparrow$       
				\Tstrut\Bstrut\\
				\hline
				\multirow{11}{*}{\rotatebox[origin=c]{90}{\textbf{$\mathbf{FT3D_{o}}$~\cite{mayer2016large}}}}
				& FlowNet3D~\cite{liu2019flownet3d}   
				& 0.212      & 0.158        & 22.86      & 58.21         
				\Tstrut\Bstrut\\
				& HPLFlowNet~\cite{gu2019hplflownet}   		
				& 0.201      & 0.169        & 26.29      & 57.45         
				\Tstrut\Bstrut\\
				& PointPWC-Net~\cite{wu2020pointpwc}  
				& 0.195      & 0.155        & 41.60      & 69.90           
				\Tstrut\Bstrut\\
				& FLOT~\cite{puy20flot}                		  
				& 0.250     & 0.153         & 39.65      & 66.08           
				\Tstrut\Bstrut\\
				& FESTA~\cite{wang2021festa}
				& -			& 0.125 		& 39.52      & 71.24  
				\Tstrut\Bstrut\\  
				& OGSFNet~\cite{ouyang2021occlusion}  
				& 0.163     & 0.122	       & 55.18	     & 77.67
				\Tstrut\Bstrut\\
				& OGSelSFNet~\cite{ouyang2021occlusion2}  
				& 0.138     & 0.103	    & 63.76			 & 82.40
				\Tstrut\Bstrut\\
				& WM3D~\cite{wang2022matters} 
				& -         & \textbf{0.063}        & \textbf{79.10}  	& \textbf{90.90}	
				\Tstrut\Bstrut\\ 
				& Bi-PointFlowNet~\cite{cheng2022bi} 
				& \textbf{0.102}     & 0.073        & \textbf{79.10}  	& 89.60							 
				\Tstrut\Bstrut\\ 
				
				& \textbf{\name{} (RS)}			
				& 0.126	 & 0.074       & 74.18	          & 88.53
				\Tstrut\Bstrut\\
				& \textbf{\name{} (FPS)}			
				& \underline{0.113} & \underline{0.066}   & \underline{77.16}	 & \underline{90.04}  
				\Tstrut\Bstrut\\
				\hdashline
				\multirow{11}{*}{\rotatebox[origin=c]{90}{\textbf{$\mathbf{KITTI_{o}}$~\cite{menze2015object}}}}
				& FlowNet3D~\cite{liu2019flownet3d}   
				& 0.175        & -    	 & 9.850         & 41.98
				\Tstrut\Bstrut\\
				& HPLFlowNet~\cite{gu2019hplflownet}   
				& 0.343        & -       & 10.30        & 38.60       
				\Tstrut\Bstrut\\
				& PointPWC-Net~\cite{wu2020pointpwc} 	
				& 0.118        & -    	 & 40.30       &  75.70
				\Tstrut\Bstrut\\
				& FLOT~\cite{puy20flot} 
				& 0.110        & -	     &41.90    	    & 72.10             		  
				\Tstrut\Bstrut\\
				& FESTA~\cite{wang2021festa}
				& 0.094	    	& - 		& 44.58      & 83.35  
				\Tstrut\Bstrut\\	 
				& OGSFNet~\cite{ouyang2021occlusion}  
				& 0.075  	    & -	   	   & 70.70   & 87.25
				\Tstrut\Bstrut\\
				& OGSelSFNet~\cite{ouyang2021occlusion2}  
				& \underline{0.060}  	    & -	   	   & 77.55  &  \underline{90.69} 
				\Tstrut\Bstrut\\
				& WM3D~\cite{wang2022matters} 
				& 0.073         & -        & \underline{81.90}  	& 89.90
				\Tstrut\Bstrut\\ 
				& Bi-PointFlowNet~\cite{cheng2022bi} 
				& 0.065     & -        & 76.90  	& 90.60							 
				\Tstrut\Bstrut\\ 
				& \textbf{\name{} (RS)}			
				& 0.067         & - 	& 80.63	         & 90.58      
				\Tstrut\Bstrut\\
				& \textbf{\name{} (FPS)}				
				& \textbf{0.051}  & -	& \textbf{89.00}  & \textbf{94.78}
				\Tstrut\Bstrut\\
			\end{tabular}
		}
	\end{center}
\end{table}

\subsection{Comparison to State-of-the-Art} \label{Quantitative}
\begin{figure*}[t]
	\begin{center}
		\includegraphics[width=1.0\linewidth]{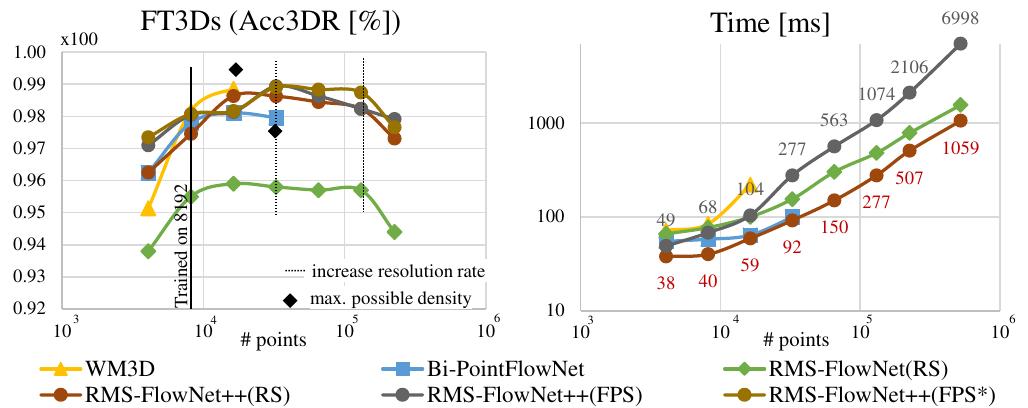}			
		\caption{Analysis of accuracy and runtime on \acrshort{ft3d} for different numbers of input points compared to state-of-the-art methods.}
		\label{Figure7_AccTimeVsMethods}
	\end{center}
\end{figure*}

To demonstrate the accuracy and generalization of our model, we compare it with state-of-the-art methods in \cref{Table2_Comparison}. 
The white cells denote the evaluation on the non-occluded $\mathrm{FT3D_s}$ as usually done in related work.
The results within the light and dark gray cells denote the evaluation with partially and extensively occluded scenes of $\mathrm{FT3D_{so}}$. 
Our \name{} allows the use of \gls{rs} and shows very comparable results to the use of \gls{fps}, but with lower runtime (\cf \cref{Figure1_Teaser}), especially for higher resolution points (\cf \cref{Figure7_AccTimeVsMethods}). 

\textbf{Evaluation on~$\mathbf{FT3D_s}$:}
We test our \name{} on non-occluded data from \gls{ft3d}, as shown in the white cells in \cref{Table2_Comparison}.
Processing the entire points with an all-to-all correlation (\ie, global correlation) using an optimal transport solver in FLOT~\cite{puy20flot} shows significantly lower accuracy than the hierarchical mechanism of our~\name{} using \gls{rs}. 
This confirms our decision to design our model in a hierarchical way.
The \gls{rs} version of our~\name{} significantly outperforms the regular representative methods as in \cite{gu2019hplflownet, gojcic2021weakly, li2021sctn}. 
This supports our decision to handle the raw points without intermediate representations for scene flow estimation. 
Furthermore, our \name{} with \gls{rs} outperforms GRU-based methods ~\cite{wei2020pv, kittenplon2021flowstep3d, gu2022rcp} with a lower runtime (\cf \cref{Figure1_Teaser}).

Compared to hierarchical designs that basically use \gls{fps}, our~\name{} with \gls{rs} outperforms ~\cite{liu2019flownet3d, wu2020pointpwc, li2021hcrf, wang2021hierarchical} on all metrics and is highly competitive with very recent methods ~\cite{cheng2022bi, wang2022matters, wang2022residual} at lower runtime as shown in \cref{Figure1_Teaser}.
However, the \gls{fps} version of our~\name{} outperforms the methods of~\cite{cheng2022bi, wang2022residual} and shows very comparable results to ~\cite{wang2022matters} with slight differences. 

Moreover, our improvements in \name{} are significant in both \gls{rs} and \gls{fps} sampling versions, even for a small number of correspondences (\ie, $K_p$ is set to $20$ with \gls{rs} and $16$ with \gls{fps}), compared to our preliminary work RMS-FlowNet~\cite{battrawy2022rms}, which uses a correspondence set of $33$ points.

\textbf{Generalization to~$\mathbf{KITTI_s}$:} 
We test the generalization ability to the KITTI data set~\cite{menze2015object} without fine-tuning, as shown in \cref{Table2_Comparison}, where the white cells denote the scores on $\mathrm{KITTI_s}$. 
Our \name{} shows a stronger generalization ability with both sampling techniques, \gls{rs} and \gls{fps}, than all state-of-the-art methods.
This is best indicated by the much smaller gap in scores between the synthetic $\mathrm{FT3D_s}$ and the realistic $\mathrm{KITTI_s}$ results.
With both sampling techniques, our~\name{} outperforms all the methods of~\cite{liu2019flownet3d, wu2020pointpwc, wang2021hierarchical, gu2019hplflownet, puy20flot, kittenplon2021flowstep3d, wei2020pv, gu2022rcp, wang2022residual}.
Compared to the competing methods in \cite{cheng2022bi, wang2022matters}, \gls{rs} with our \name{} shows comparable results, but with lower runtime (\cf \cref{Figure1_Teaser}), and \gls{fps} outperforms these methods for all metrics.

\textbf{Robustness to Occlusions:} 
Training with non-occluded points using $\mathrm{FT3D_s}$ shows that our \name{} is able to estimate a reasonable accuracy of scene flow on the test split data when evaluated on $\mathrm{FT3D_{so}}$, as shown in the light and dark gray cells in \cref{Table2_Comparison}. 
When evaluated on the $\mathrm{FT3D_{so}}$, the scores of all input points are reported, taking into account the partial or large number of occluded points.
In the evaluation of the $\mathrm{FT3D_{so}}$ test split with occlusions, the \gls{fps} and \gls{rs} versions of our \name{} take second and third place, respectively, behind the method in \cite{wang2022matters}.

On $\mathrm{KITTI_d}$, \ie including occluded points in the evaluation, the \gls{fps} version of our \name{} shows the best results of all methods in all metrics, and the faster \gls{rs} version of our method shows comparable results to competing methods \cite{cheng2022bi, wang2022matters} when evaluated.

\Cref{Table3_Occlusions} shows the results on $\mathrm{FT3D_o}$ and $\mathrm{KITTI_o}$, where occluded points are additionally considered during training.
For a fair comparison, we follow the other method's evaluation scheme and include the occlusions during inference and evaluate over all input points in the EPE3D metric. 
In the $\mathrm{EPE3D_{noc}}$, $\mathrm{Acc3DS_{noc}}$, and $\mathrm{Acc3DR_{noc}}$ metrics, we ignore the occluded points when computing the scores, but still include them as input.
For both data sets, $\mathrm{FT3D_o}$ and $\mathrm{KITTI_o}$, our \name{} with \gls{rs} ranks right behind the competing methods \cite{cheng2022bi, wang2022matters}, but with \gls{fps} we rank second for $\mathrm{FT3D_o}$ and first for $\mathrm{KITTI_o}$.

\subsection{Varying Point Densities} \label{Density}
We evaluate the two versions of our \name{}, \ie with \gls{rs} and \gls{fps}, against our earlier work \cite{battrawy2022rms} and the competing methods \cite{cheng2022bi, wang2022matters} in terms of accuracy (Acc3DR) and runtime at different input densities.
The results of this comparison are shown in \cref{Figure7_AccTimeVsMethods}.
We consider a wide range of densities ${N = \{4096 * 2^i\}^7_{i=0}}$ of $\mathrm{FT3D_s}$, and finally all available non-occluded points are evaluated, which corresponds to \mytilde$225$K points on average.
For a fair comparison, all methods are trained with a fixed resolution of $8192$ points only, and we do not consider fine-tuning or further training with different point densities.
We measure the inference time for all methods equally on a Geforce GTX 1080 Ti, including our \name{} and RMS-FlowNet \cite{battrawy2022rms} with \gls{rs}.

For both \gls{rs} and \gls{fps} versions of our method, to keep the accuracy stable for densities $>32$K, we increase the resolution rate of the downsampling layers (\cf \cref{feature_extraction}) to ${\{\{l\}}^3_{k=1}\mid l_1 = 4096, l_2 = 1024, l_3 = 256\}$ without any further training or fine-tuning.
Furthermore, for the \gls{rs} of our \name{} for densities $>131$K, we increase the resolution rate of the downsampling layers to ${\{l\}}^3_{k=1}\mid l_1=8192, l_2=2048, l_3=512\}$ without further training or fine-tuning. 
Increasing the resolution of the downsampling layers is not possible with \gls{fps}, as this would exceed the memory limit of the Geforce GTX 1080 Ti.
Nevertheless, the accuracy of our method remains stable over a wide range of densities for both \gls{rs} and \gls{fps} versions (\cf \cref{Figure7_AccTimeVsMethods}).
\begin{figure*}[t]
	\begin{center}
		\includegraphics[width=1.0\linewidth]{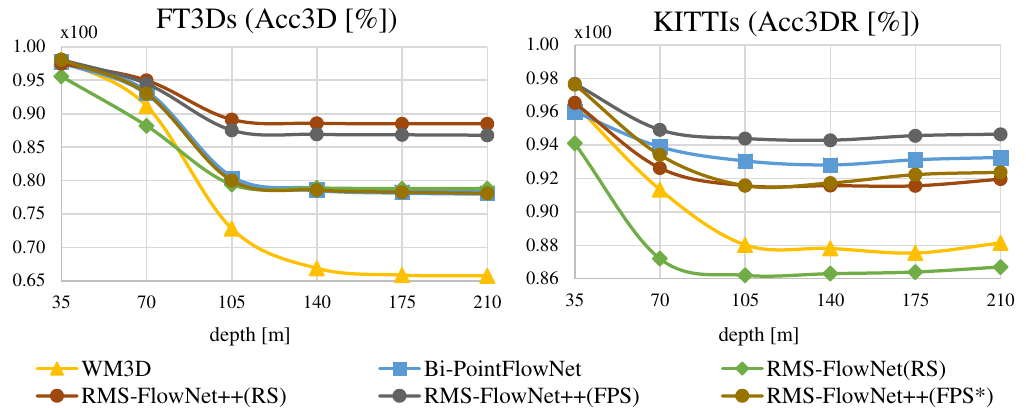}			
		\caption{Analysis of accuracy for different depth limits on $\mathrm{FT3D_s}$ and $\mathrm{KITTI_s}$ compared to state-of-the-art methods.}
		\label{Figure8_DepthRangesVsMethods}
	\end{center}
\end{figure*}
To maintain the accuracy of WM3D \cite{wang2022matters} and to evaluate more than $8192$ input points, we had to increase the resolution rate of the downsampling layers based on the resolutions suggested in \cite{wang2022matters}. 
Yet, for the competing methods WM3D~\cite{wang2022matters} and Bi-PointFlowNet \cite{cheng2022bi}, the maximum possible densities are limited to $16384$ and $32768$, respectively, since they exceed the memory limit of the Geforce GTX 1080 Ti at higher densities.
For the other state-of-the-art-methods FLOT~\cite{puy20flot}, PV-RAFT~\cite{wei2020pv}, PointPWC-Net~\cite{wu2020pointpwc}, and HPLFlowNet~\cite{gu2019hplflownet}, the maximum possible densities are limited to $8192$, $8192$, $32768$ and $65536$, respectively, for the same reason (not shown in \cref{Figure7_AccTimeVsMethods}).

In contrast, our \name{} allows very high densities with high accuracy without exceeding the memory limit of the Geforce GTX 1080 Ti.
Although \gls{fps} is computationally expensive, the reduced number of nearest neighbors ($K_p=16$) allows the operation with \mytilde$225$K points.
Using \gls{rs} with the increased number of nearest neighbors ($K_p=20$) allows our \name{} to operate 5 to 6 times faster than with \gls{fps}, especially at densities $>65K$. 
Consequently, the design of~\name{} allows a much higher maximum density compared to other methods in terms of memory requirement and time consumption.
However, the runtime of our~\name{} still increases super-linearly with increasing input density $>225$K due to the~\acrshort{knn} search. 
In addition, the initial drop in accuracy at \mytilde$225$K points in \cref{Figure7_AccTimeVsMethods}  indicates that it may be necessary to further increase the resolution of the downsampling layers when using even higher densities.

\begin{figure*}[t]
	\begin{tabular}{p{0.1cm}cccc}
		& \textit{Example 1}  & \textit{Example 2} & \textit{Example 3} 
		
		\Tstrut\Bstrut\\ 
		\rotatebox[origin=c]{90}{\textit{Scene}} &
		\includegraphics[width=0.3\linewidth,valign=c]{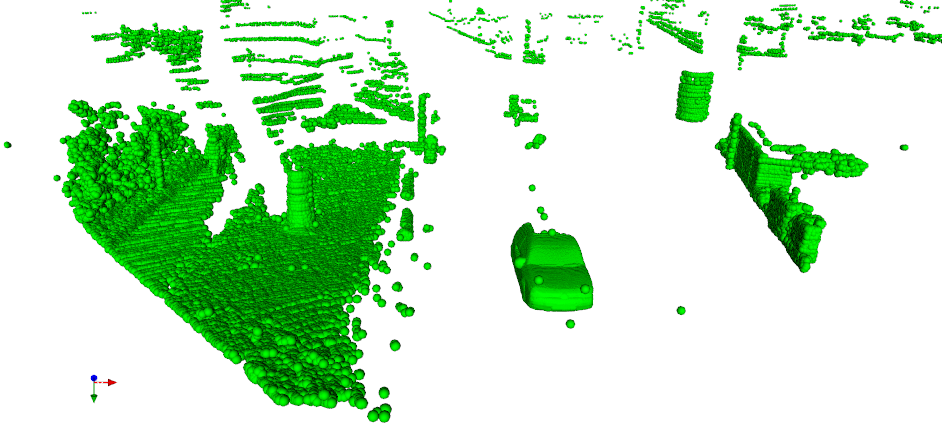}&
		\includegraphics[width=0.3\linewidth,valign=c]{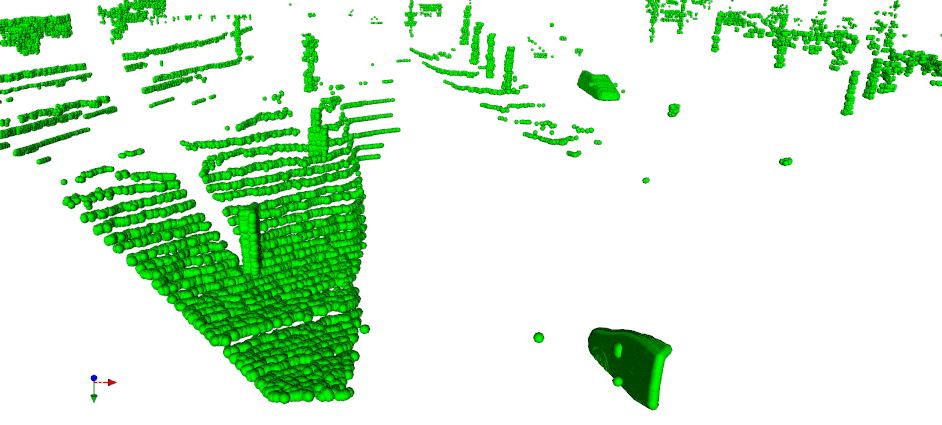}&
		\includegraphics[width=0.3\linewidth,valign=c]{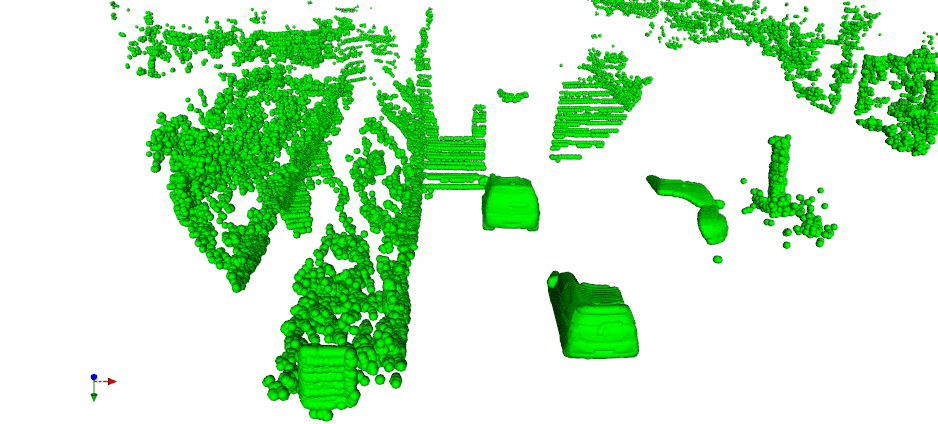}
		\Tstrut\Bstrut\\	
		
		\rotatebox[origin=c]{90}{\textit{Error (210m)}} &
		\includegraphics[width=0.3\linewidth,valign=c]{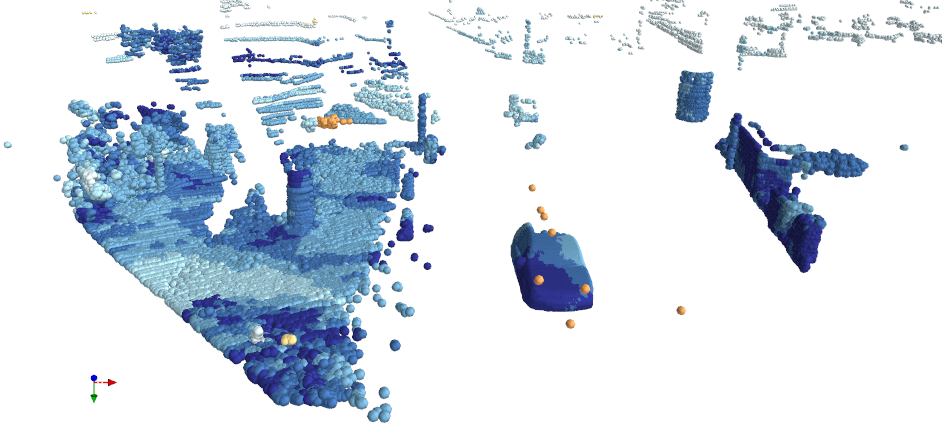}&
		\includegraphics[width=0.3\linewidth,valign=c]{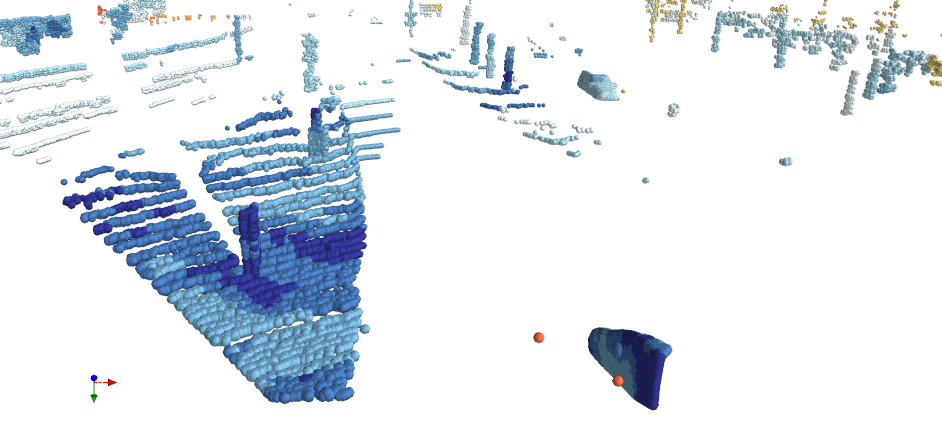}&
		\includegraphics[width=0.3\linewidth,valign=c]{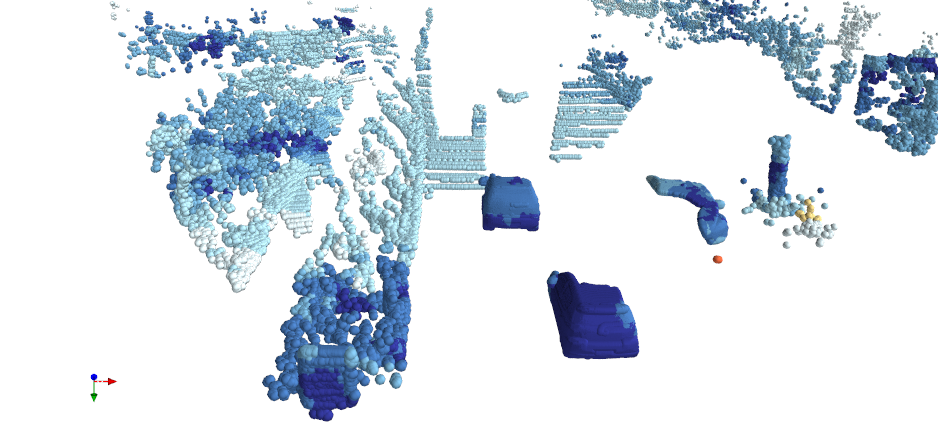}
		\Tstrut\Bstrut\\
		
		\Tstrut\Bstrut\\ 
		\rotatebox[origin=c]{90}{\textit{Scene}} &
		\includegraphics[width=0.3\linewidth,valign=c]{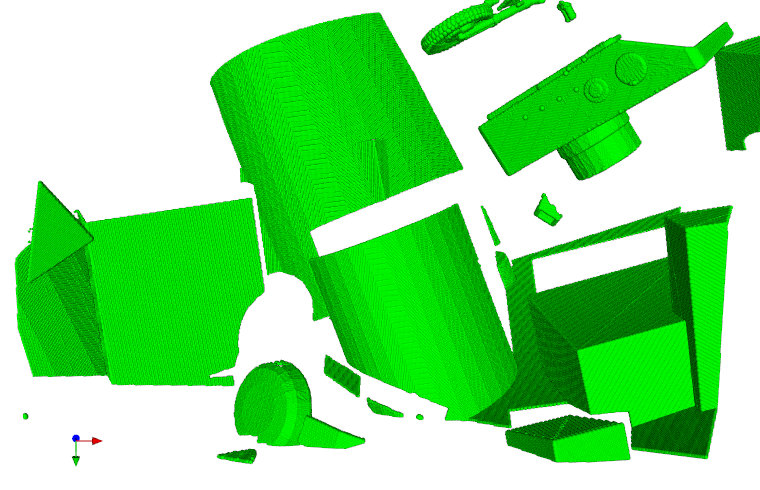}&
		\includegraphics[width=0.3\linewidth,valign=c]{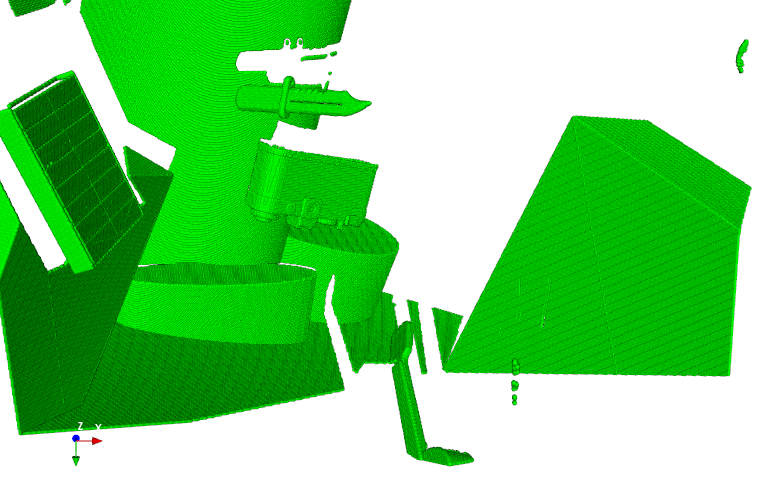}&
		\includegraphics[width=0.3\linewidth,valign=c]{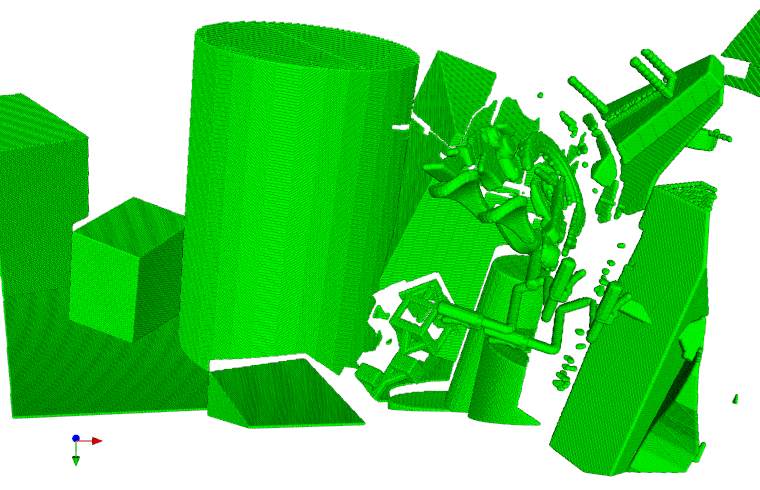}
		\Tstrut\Bstrut\\	
		
		\rotatebox[origin=c]{90}{\textit{Error (35m)}} &
		\includegraphics[width=0.3\linewidth,valign=c]{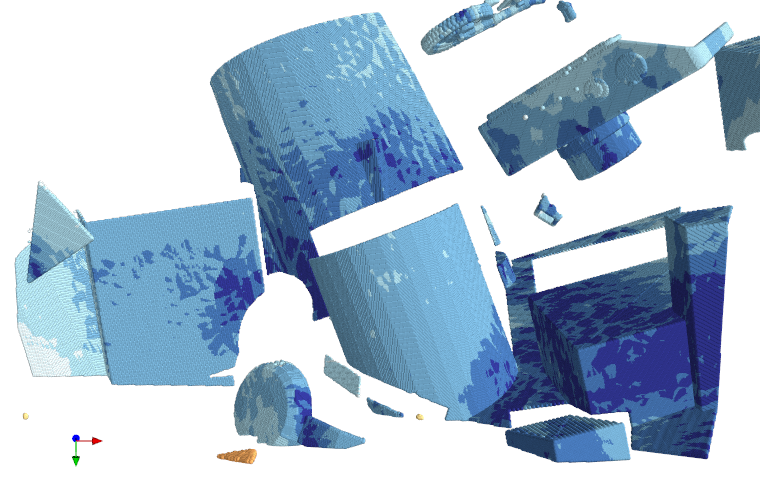}&
		\includegraphics[width=0.3\linewidth,valign=c]{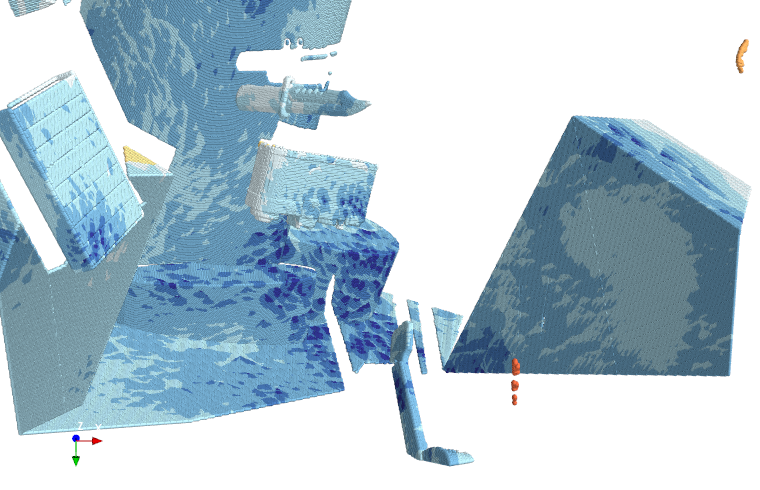}&
		\includegraphics[width=0.3\linewidth,valign=c]{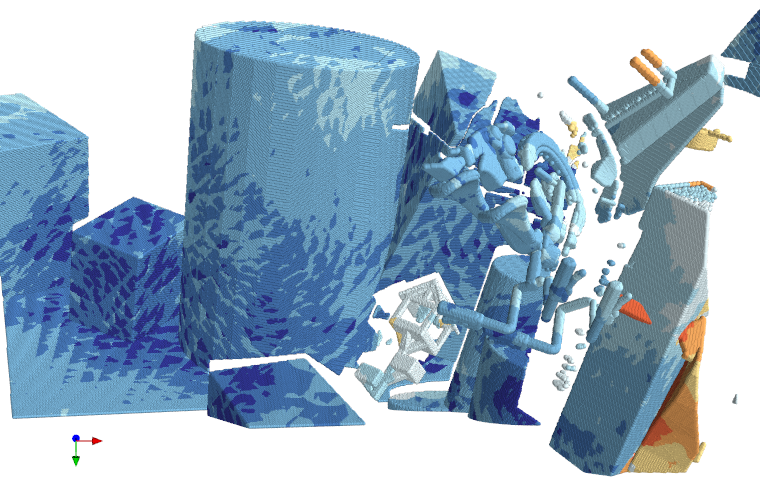}
		\Tstrut\Bstrut\\
		& \multicolumn{3}{c}{\includegraphics[width=0.95\linewidth,valign=c]{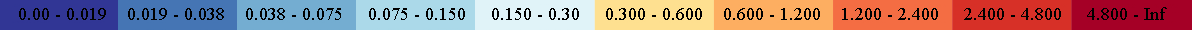}}\Tstrut\Bstrut\\
	\end{tabular}
	\caption{Three examples from the non-occluded versions of $\mathrm{FT3D_s}$ and $\mathrm{KITTI_s}$ show that our \name{} allows high point densities with high accuracy using \gls{rs}. The scene of each example (first and third rows) visualizes $P^t$ as green color. The error map of each scene (second and forth rows) shows the end-point error in meters according to the color map shown in the last row.}
	\label{Figure9_QualitativeDensity}
\end{figure*}

We visually present some results on $\mathrm{KITTI_s}$ with dense points (\mytilde $50$K points) and three examples of non-occluded points of the $\mathrm{FT3D_s}$ (\mytilde $300$K points) in \cref{Figure9_QualitativeDensity}.
To obtain a denser scene in the $\mathrm{KITTI_s}$, we include distant points down to $<210$ meters.
Our \name{} shows a high accuracy even with this very dense data.

\subsection{Varying Depth Ranges} \label{Depth}
We emphasize that all state-of-the-art methods only consider objects in the near range ($<35$ meters) during training and evaluation. 
We consider the same range during training, but in this work, for the first time, we evaluate the accuracy of scene flow for more distant objects using the $\mathrm{FT3D_s}$ and $\mathrm{KITTI_s}$ data sets (\cf~\cref{Figure8_DepthRangesVsMethods}).

\begin{figure*}[t]
	\begin{tabular}{p{0.1cm}cccc}
		& \textit{Bi-PointFlowNet~\cite{cheng2022bi}} & \textbf{\textit{Our \name{} (RS)}} & \textbf{\textit{Our \name{} (FPS)}} 
		\Tstrut\Bstrut\\ 
		\rotatebox[origin=c]{90}{\textit{Scene + SF}} &
		\includegraphics[width=0.3\linewidth,valign=c]{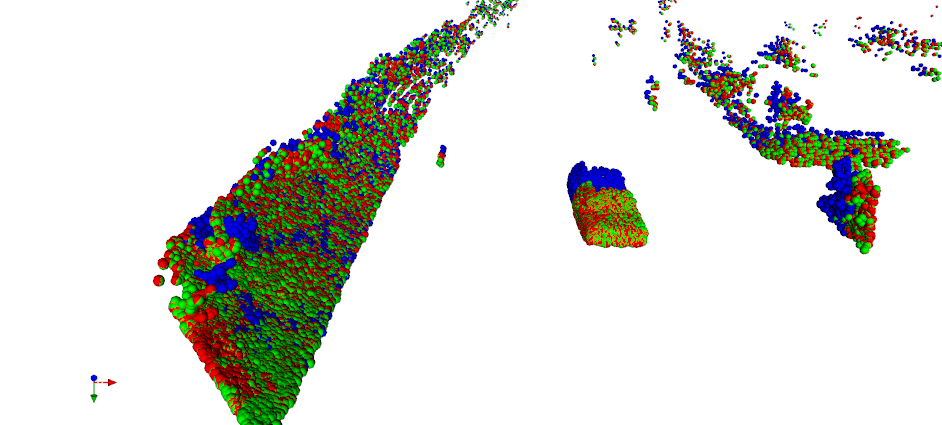}&
		\includegraphics[width=0.3\linewidth,valign=c]{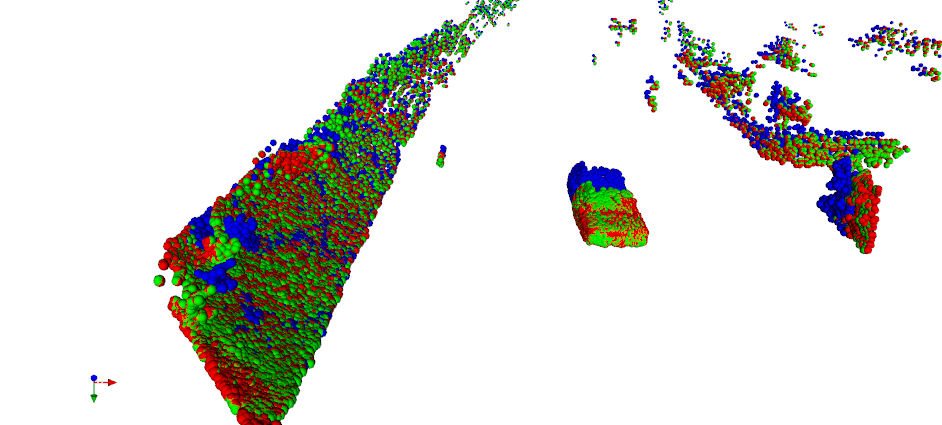}&
		\includegraphics[width=0.3\linewidth,valign=c]{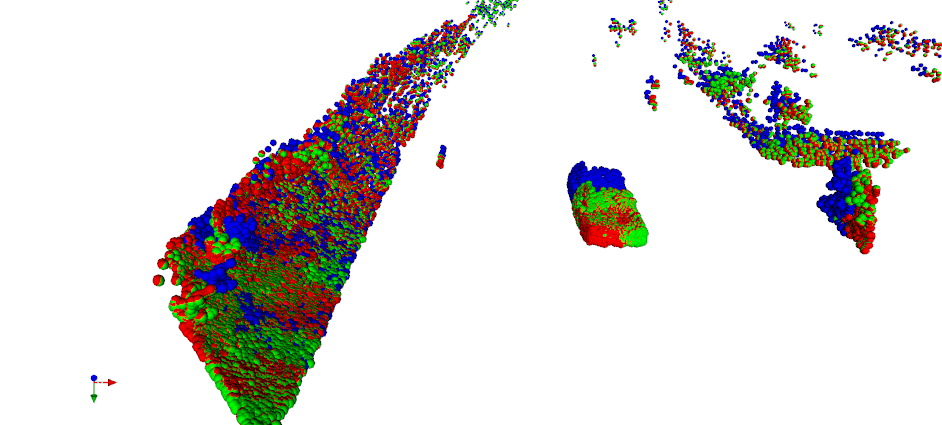}
		\Tstrut\Bstrut\\	
		
		\rotatebox[origin=c]{90}{\textit{Error (35m)}} &
		\includegraphics[width=0.3\linewidth,valign=c]{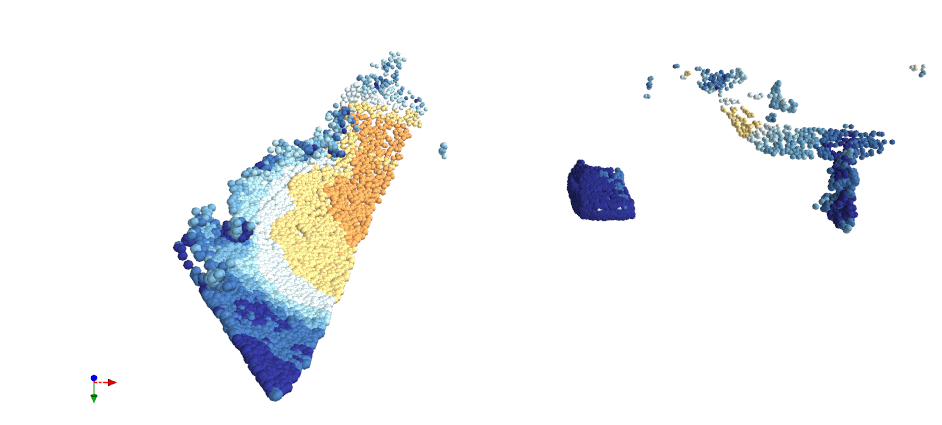}&
		\includegraphics[width=0.3\linewidth,valign=c]{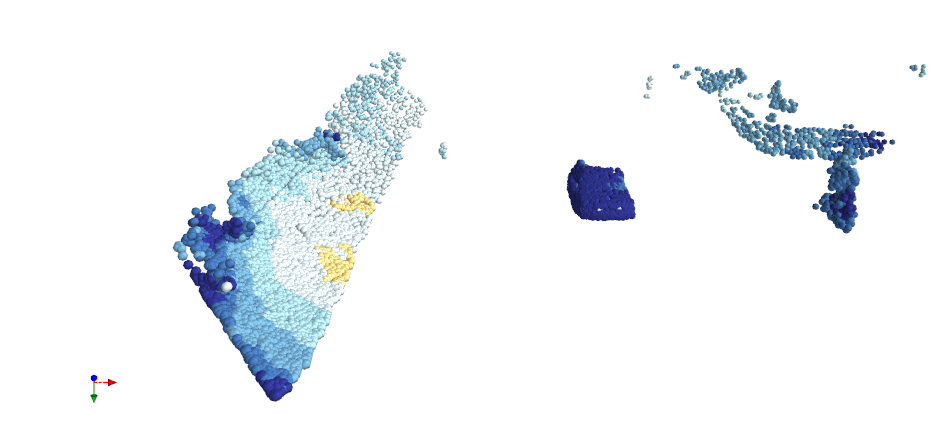}&
		\includegraphics[width=0.3\linewidth,valign=c]{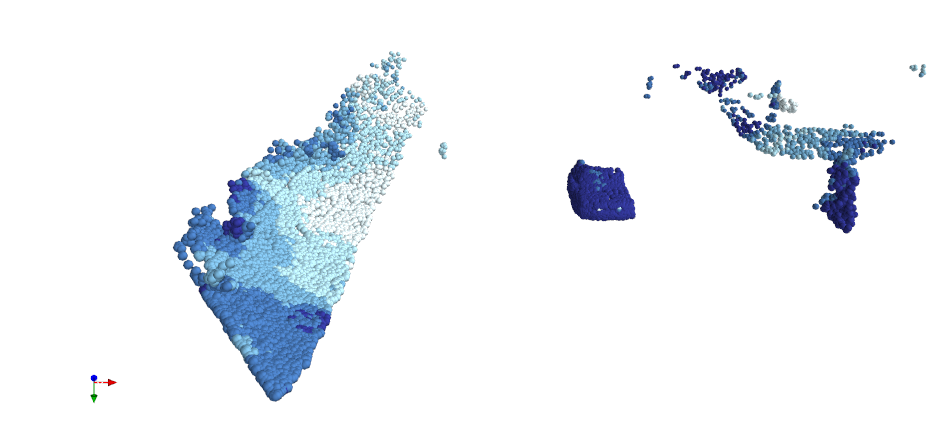}
		\Tstrut\Bstrut\\
		
		\rotatebox[origin=c]{90}{\textit{Error (210m)}} &
		\includegraphics[width=0.3\linewidth,valign=c]{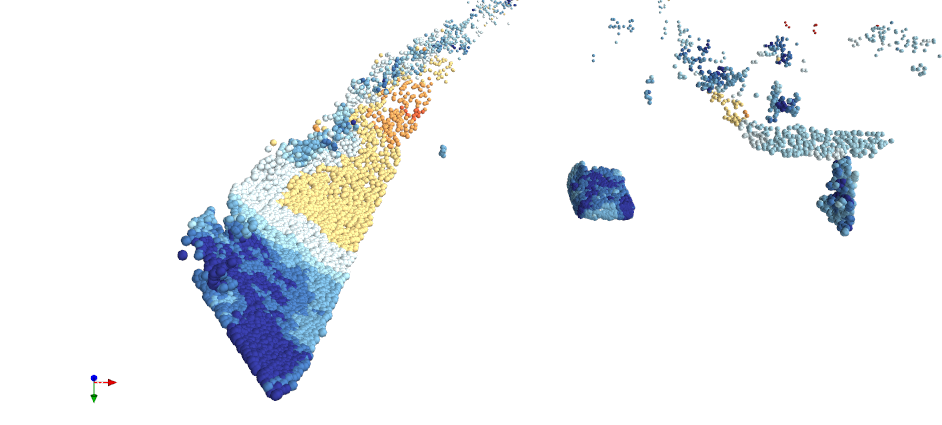}&
		\includegraphics[width=0.3\linewidth,valign=c]{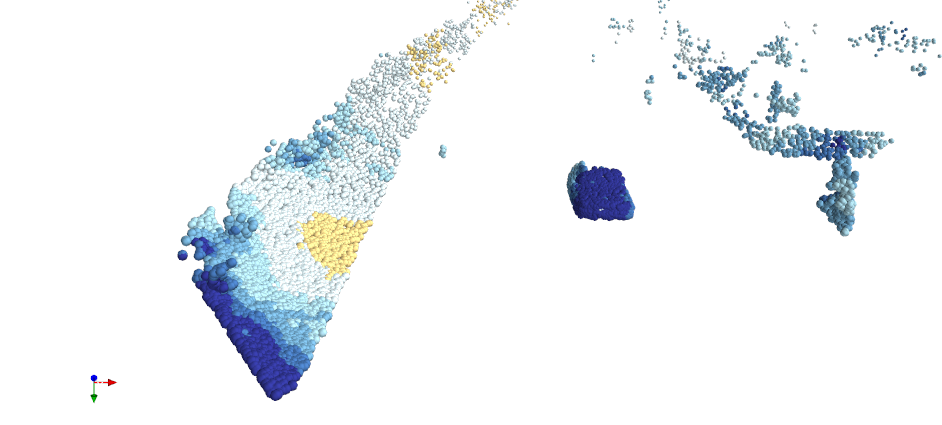}&
		\includegraphics[width=0.3\linewidth,valign=c]{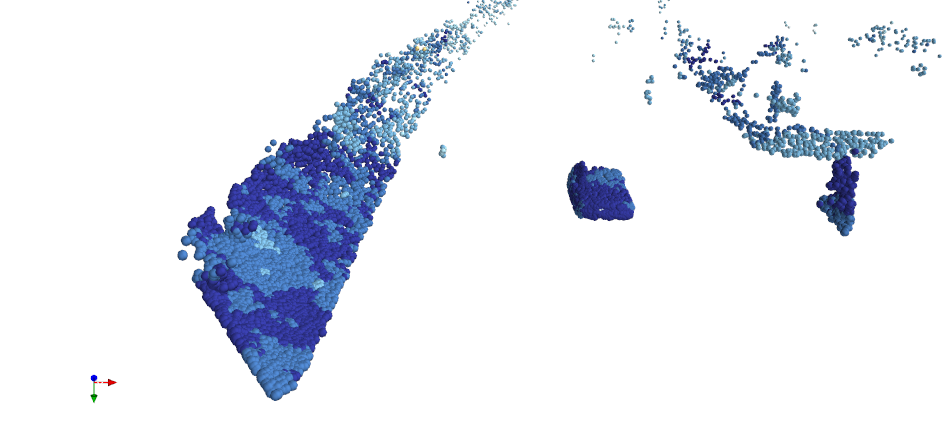}
		\Tstrut\Bstrut\\	
		
		\Tstrut\Bstrut\\ 
		\rotatebox[origin=c]{90}{\textit{Scene + SF}} &
		\includegraphics[width=0.3\linewidth,valign=c]{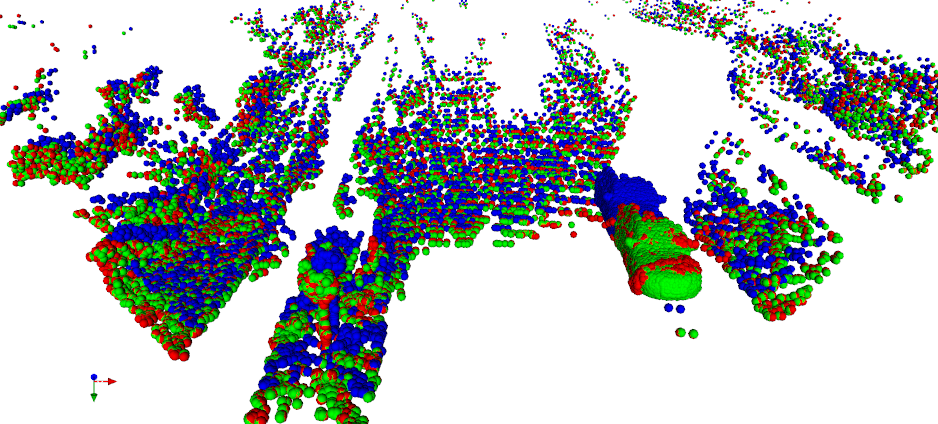}&
		\includegraphics[width=0.3\linewidth,valign=c]{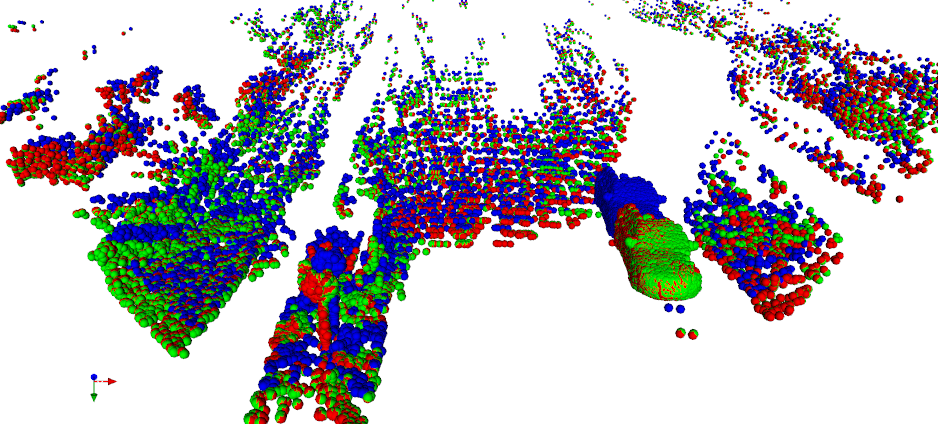}&
		\includegraphics[width=0.3\linewidth,valign=c]{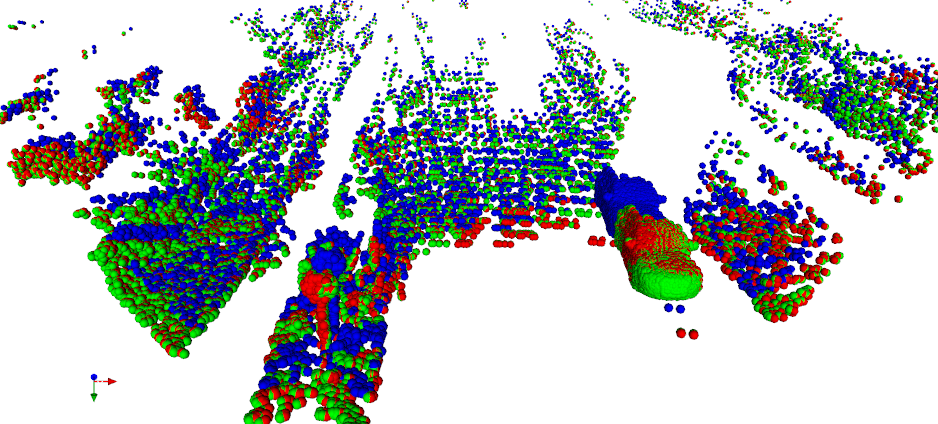}
		\Tstrut\Bstrut\\	
		
		\rotatebox[origin=c]{90}{\textit{Error (35m)}} &
		\includegraphics[width=0.3\linewidth,valign=c]{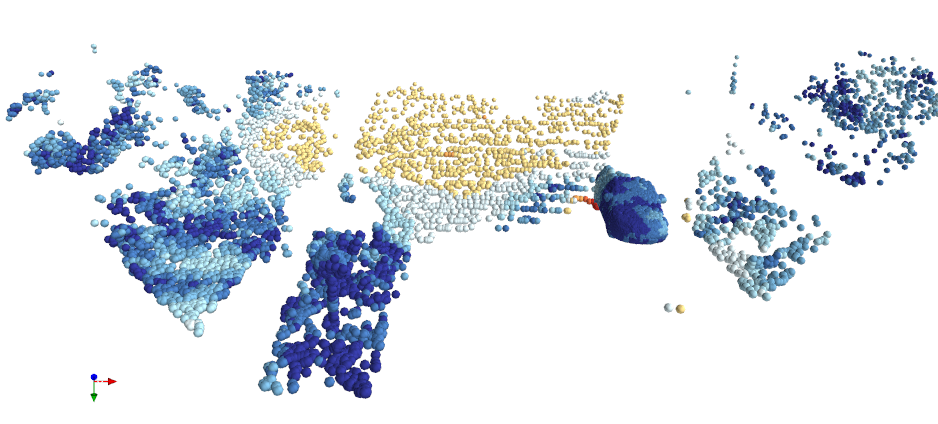}&
		\includegraphics[width=0.3\linewidth,valign=c]{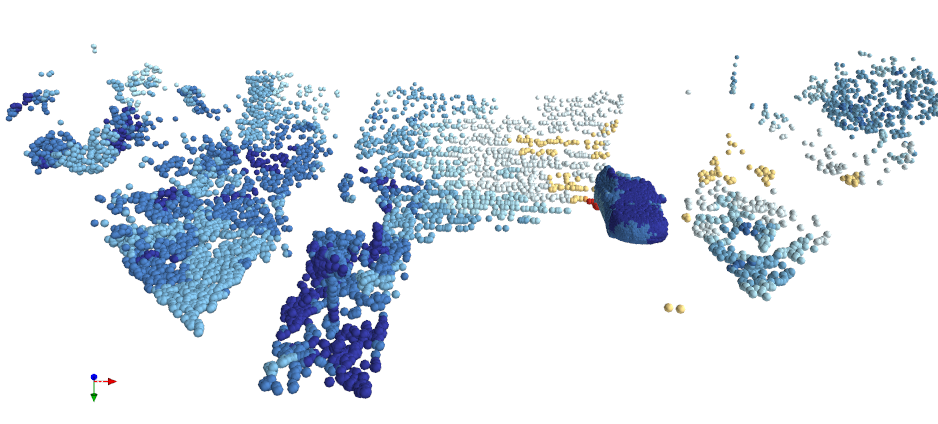}&
		\includegraphics[width=0.3\linewidth,valign=c]{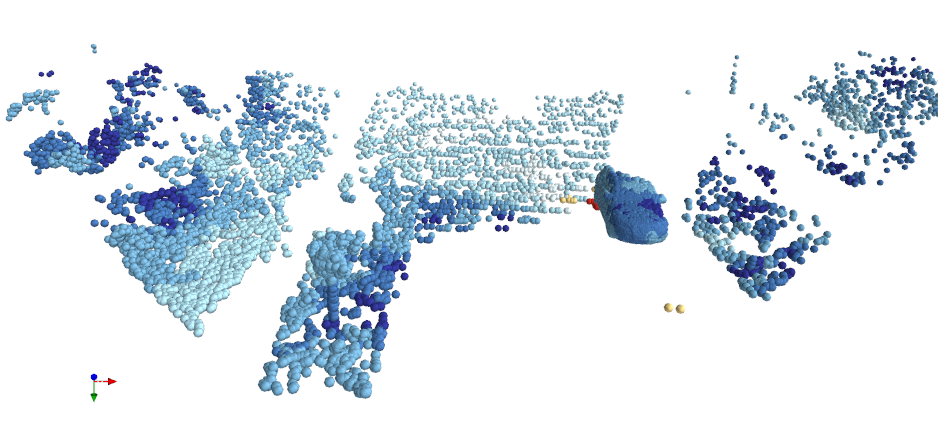}
		\Tstrut\Bstrut\\
		
		\rotatebox[origin=c]{90}{\textit{Error (210m)}} &
		\includegraphics[width=0.3\linewidth,valign=c]{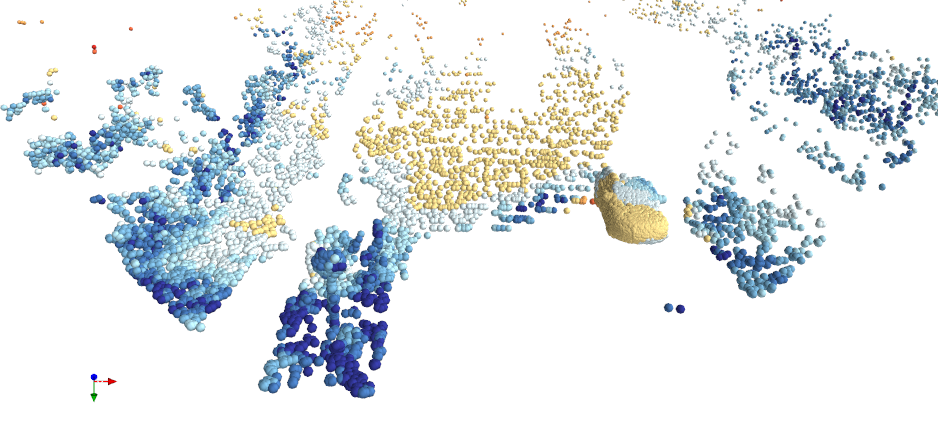}&
		\includegraphics[width=0.3\linewidth,valign=c]{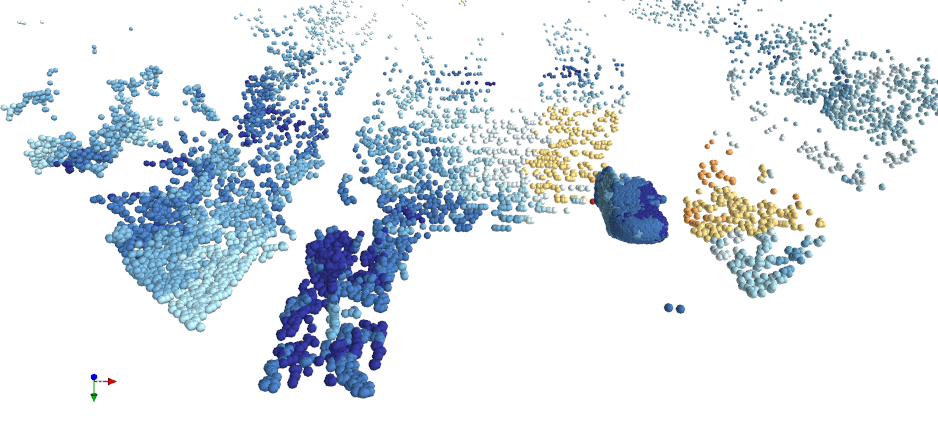}&
		\includegraphics[width=0.3\linewidth,valign=c]{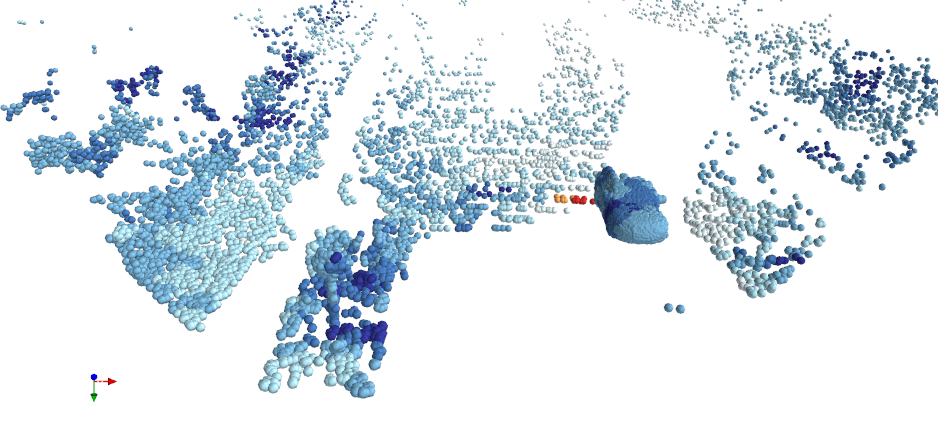}
		\Tstrut\Bstrut\\
		
		& \multicolumn{3}{c}{\includegraphics[width=0.95\linewidth,valign=c]{KITTI_errorcolors_3D.png}}\Tstrut\Bstrut\\
	\end{tabular}
	\caption{Two examples taken from $\mathrm{KITTI_s}$ show the impact of our \name{} compared to the competing method Bi-PointFlowNet~\cite{cheng2022bi}. The scene of each example (first and forth rows) visualizes $P^t$ as blue color and the predicted and ground truth scene flow after adding them to $P^t$ in green and red color, respectively. The error map of each scene (second, third, fifth and sixth rows) shows the end-point error in meters according to the color map shown in the last row. Our \name{} shows lower errors (dark blue) over a wide area of the observed scene, compared to the competing method.}
	\label{Figure10_QualitativeComparison}
\end{figure*}

On \gls{ft3d}, the accuracy of our \name{} with \gls{rs} and \gls{fps} is better than the competing methods for every depth limit. 
The accuracy of WM3D \cite{wang2022matters} decreases significantly, and the accuracy of Bi-PointFlowNet \cite{cheng2022bi} is $\mytilde8\%$ lower than ours. 
Surprisingly, \gls{rs} generalizes slightly better than \gls{fps} to increasing depth limits.
When \name{} is trained and evaluated with \gls{fps} (marked with $^*$), the results are on par with our prior work RMS-FlowNet~\cite{battrawy2022rms} and the competing method Bi-PointFlowNet \cite{cheng2022bi}.
When evaluated on $\mathrm{KITTI_s}$, our \name{} shows significantly better results than our prior work RMS-FlowNet \cite{battrawy2022rms}.
Furthermore, both sampling strategies perform significantly better than WM3D \cite{wang2022matters}.
However, when trained with \gls{rs} and evaluated with \gls{fps}, the scene flow accuracy exceeds that of Bi-PointFlowNet \cite{cheng2022bi}.

It follows that training with \gls{rs} can better generalize to a wider range of points than \gls{fps},
leading to better scene flow accuracy with \gls{fps} than training with \gls{fps} itself. 
In other words, \gls{fps} causes the downsampled points to cover approximately the same spatial locations, which reduces the variation during training.

Qualitatively, we visualize two scenes of $\mathrm{KITTI_s}$ with the corresponding error maps in \cref{Figure10_QualitativeComparison}.
We qualitatively compare our \name{} with both sampling techniques to the most competitive method~\cite{cheng2022bi}.
Without changing the training strategy, we compare the predicted scene flow with the narrowest depth range ($<35$m) and with the widest range ($<210$m) against Bi-PointFlowNet.
Testing within the trained depth range ($<35$m), we see that Bi-PointFlowNet has higher errors on flat surfaces than our approach, which produces the best results with \gls{fps}. 
Testing outside the trained depth range ($>35$m) shows that the accuracy decreases when distant points ($<210$m) are included.
In \cite{cheng2022bi}, even nearby objects (\eg moving cars) are negatively affected when distant points ($>35$m) are included in the scene.
However, our \name{} performs robustly in this case.

\subsection{Ablation Study} \label{Ablation}
\begin{table*}[t]
	\caption{We explore our improvements in \name{} compared to our preliminary work RMS-FlowNet~\cite{battrawy2022rms}. To speed up the experiments, we use \gls{fps} during training and evaluation. The first line corresponds to the method in \cite{battrawy2022rms}.}
	\label{Table4_Ablation}
	\begin{center}
		\resizebox{1.0\linewidth}{!}
		{
			\begin{tabular}{ccccc|cc|cc}
				\multirow{3}{*}{$\mathbf{K_p}$}
				& \multirow{3}{*}{\textbf{Graph Rep.}}				
				& \multirow{3}{*}{\textbf{Decoder}} 
				& \multirow{3}{*}{\textbf{$2^{nd}$ Embedding}}
				& \multirow{3}{*}{\textbf{Bidirectional Map}}
				& \multicolumn{2}{c|}{\textbf{\acrshort{ft3d_bold}~\cite{mayer2016large}}} 
				& \multicolumn{2}{c}{$\mathbf{KITTI_{s}}$\textbf{\cite{menze2015object}}}
				\\
				& & &   &  		& \textbf{EPE3D}  & \textbf{Acc3DR} & \textbf{EPE3D} & \textbf{Acc3DR} \\	
				& & &   &  		& [m]             & [\%]            & [m]             & [\%] 
				\Tstrut\Bstrut\\
				\hline			
				33 & \xmark & \cmark  & \xmark  & \xmark   & 0.051           & 95.60        & 0.047          & 95.80            
				\Tstrut\Bstrut\\		
				16 & \xmark & \cmark  & \xmark  & \xmark    & 0.054           & 95.15         & 0.067          & 91.72            
				\Tstrut\Bstrut\\
				16 & \cmark & \cmark & \xmark  & \xmark     & 0.041           & 96.85         & 0.043          & 94.16      
				\Tstrut\Bstrut\\
				16 & \cmark & \xmark & \xmark  & \xmark     & 0.034          & 97.63         & 0.039         & 94.57         
				\Tstrut\Bstrut\\
				16 & \cmark & \xmark & \cmark  & \xmark    & 0.033          & 97.77         & 0.036         & 95.52         
				\Tstrut\Bstrut\\
				16 & \cmark & \xmark  & \cmark  & \cmark    & \textbf{0.029}          & \textbf{98.09}         & \textbf{0.030}         & \textbf{97.63}      
				\Tstrut\Bstrut\\
				\hdashline
				16 &   \multicolumn{4}{c|}{\textbf{Trained with \gls{rs} with $\mathbf{K_p=20}$}}  & \textbf{0.029}      & \textbf{98.10}         & \textbf{0.027}         & \textbf{97.67}     
			\end{tabular}
		}
	\end{center}
\end{table*}
To speed up our experiments, we verify our design decisions, the additional components in the \gls{fe} by training with \gls{fps}, which converges faster than \gls{rs}. 
We also compare \gls{rs} with \gls{fps} and increase the \glspl{knn} to test the effect on the results. 
Finally, we compare the impact of our augmentation on the overall results with \gls{rs}. 

\textbf{Design Decisions:}
First, we reduce the number of correlation points ($K_p$) from $33$ in our preliminary work of RMS-FlowNet \cite{battrawy2022rms} to $16$ in the whole design (\ie, all scales of \gls{lfa} and \gls{fe}), then we verify our improvements in \name{} as shown in \cref{Table4_Ablation}.
The accuracy of our preliminary work RMS-FlowNet decreases by reducing the number of correlation points to $16$, but it is improved again by using the graph representation of \cref{Equation1_GraphRepresentation} in our \gls{fe}, which adds the $f_{i}^{t}$ part to the original representation in RMS-FlowNet \cite{battrawy2022rms}.
Second, we slightly improve the results by omitting the decoder part in the feature extraction which saves more operations and upsampling layers and avoids the use of \gls{knn} search.
Third, we show the positive effect of adding the $2^{nd}$ embedding step to the \gls{fe} of RMS-FlowNet \cite{battrawy2022rms}, which is based on the feature space.
Then, we check the positive effect of our similarity map (\ie, one-to-one map) based on the feature space to find a pair of matching points under a bidirectional constraint as explained in \cref{Bidirectional_Map}.
Finally, we show the final results using the \gls{fps} of the model, but trained using the \gls{rs} sampling technique with correspondence set ($K_{p}$) during training. 
Compared to RMS-FlowNet \cite{battrawy2022rms}, the reduction of $K_{p}$ makes the method more efficient, while the sum of architectural changes improves the results. 

\textbf{Aspects of Attention in \gls{fe}:}
We design our \acrfull{fe} with two maximum embedding layers based on both Euclidean and feature space followed by two attentive embedding layers (see \cref{Figure6_FlowEmbedding}). 
Focusing on our stacked attention layers, we verify three important aspects of our stacked attention design on the \gls{ft3d} data set as follows:
\begin{enumerate}
	\item Adding the feature of reference frame $f_{i}^{t}$ (as in \cref{Equation5_3rdEmbedding}) to the input of the attention mechanism. 
	\item Adding the residual connection (Res. Conn.) as in \Cref{Figure6_FlowEmbedding} or $e_{2i}^{t}$ as in \cref{Equation10_residual}.
	\item Encoding of the spatial locations to the features and the concatenation to the $\hat{f}_{i}^{t}$ in the \cref{Equation5_3rdEmbedding}. 
\end{enumerate}
Combining all of the above components yields more accurate results, as verified in \cref{ablation}. 

\begin{table}[t]
	\caption{We verify the aspects of our stacked attention in flow embedding on \gls{ft3d} data set. To speed up the experiments, we use \gls{fps} for training and evaluation.}
	\label{ablation}
	\begin{center}
		\resizebox{1.0\linewidth}{!}
		{
			\begin{tabular}{ccc|cc}
				\\
				\textbf{Spatial} & \multirow{2}{*}{\textbf{$\mathbf{f_{i}^{t}}$}} & \textbf{Res.} &   \textbf{EPE3D}  & \textbf{Acc3DR} 
				\\	
				\textbf{Encoding} & &  \textbf{Conn.} &   [m] $\downarrow$ & [\%] $\uparrow$ 
				\Tstrut\Bstrut\\
				\hline	
				\xmark & \xmark  & \xmark     & 0.050          & 95.13         
				\Tstrut\Bstrut\\			
				\cmark & \xmark  & \xmark     & 0.040          & 96.84         
				\Tstrut\Bstrut\\		
				\cmark & \cmark  & \xmark     & 0.038          & 97.20        
				\Tstrut\Bstrut\\
				\xmark & \cmark & \cmark      & 0.038          & 96.96         
				\Tstrut\Bstrut\\
				\cmark & \cmark & \cmark     & \textbf{0.029}          & \textbf{98.09}         
				\Bstrut\\
				\hdashline
				\multicolumn{3}{c|}{\textbf{Trained with \gls{rs} with $\mathbf{K_p=20}$}}  & \textbf{0.029}      & \textbf{98.10}        
				\Tstrut  
			\end{tabular}
		}
	\end{center}
\end{table}

\textbf{Training with \gls{fps} vs. \gls{rs}:}
First, we train with \gls{fps} using different numbers of $K_{p}$ and evaluate with the same numbers used in training to determine the appropriate number of $K_{p}$ (\ie, the correct correspondence set) that gives the best results, as shown in \cref{Table5_Ablation_NumberKNN}.
We find that small correspondence sets such as 8 and 12 have lower accuracy than 16 and 20, which both give roughly comparable results, making them appropriate numbers, but at the cost of a higher number of FLOPs. 
Based on this, we train with \gls{rs} and these determined numbers of $K_{p}$ (\ie, 16 and 20).
After evaluation, we find that $K_{p}=20$ works best with \gls{rs}, as shown in \cref{Table5_Ablation_NumberKNN}.
Based on this, we set $K_p$ to 16 and 20 for \gls{fps} and \gls{rs}, respectively. 
We then perform cross evaluations with both sampling techniques to verify which sampling method generalizes better.
The results are shown in \cref{Table6_Ablation_FPSvsRS}.
\Gls{rs} generalizes much better than \gls{fps}.
It even improves the results when evaluated with \gls{fps}, compared to the original training with \gls{fps}.

\begin{table}[b]
	\caption{We evaluate the number of \glspl{knn} that can be used for both \gls{fps} and \gls{rs} sampling techniques. Training and evaluation are always performed with the same sampling technique and number of nearest neighbors. FLOPs do not count for the sampling or for the nearest neighbor search.}
	
	\label{Table5_Ablation_NumberKNN}
	\begin{center}
		\resizebox{1.0\linewidth}{!}
		{
			\begin{tabular}{c|c|cc|cc|c}				
				\textbf{Sampling}
				& \multirow{3}{*}{$\mathbf{K_p}$} 
				& \multicolumn{2}{c|}{$\mathbf{FT3D_{s}}$~\textbf{\cite{mayer2016large}}} 
				& \multicolumn{2}{c}{$\mathbf{KITTI_{s}}$\textbf{\cite{menze2015object}}}\\
				\textbf{Technique}  &          		& \textbf{EPE3D}  & \textbf{Acc3DR} & \textbf{EPE3D} & \textbf{Acc3DR} & \textbf{FLOPs}\\	
				&          		& [m]             & [\%]            & [m]             & [\%]           & [G] 
				\Tstrut\Bstrut\\
				\hline			
				\multirow{4}{*}{\textbf{FPS}} 
				& 8       & 0.035             & 97.70             & 0.043              & 95.38             & \textbf{17.16}       
				\Tstrut\Bstrut\\
				& 12      & 0.031             & 98.04             & 0.031              & 97.54             & 23.55         
				\Tstrut\Bstrut\\		
				& \textbf{16}      & \textbf{0.029}    & \textbf{98.09}    & 0.030              & \textbf{97.63}    & 29.93      
				\Tstrut\Bstrut\\
				& \textbf{20}      & \textbf{0.029}    & 98.07             & \textbf{0.029}     & 97.56             & 36.32      
				\Tstrut\Bstrut\\
				\hdashline
				\multirow{2}{*}{\textbf{RS}} 
				& 16      & 0.034             & 97.50             & 0.047              & 94.21             & \textbf{29.93}
				\Tstrut\Bstrut\\
				& \textbf{20}      & \textbf{0.033}    & \textbf{97.51}   & \textbf{0.035}      & \textbf{96.53}    & 36.32
				\Tstrut\Bstrut\\
			\end{tabular}
		}
	\end{center}
\end{table}

\begin{table}[b]
	\caption{We evaluate the generalization of each sampling technique to the other on $\mathrm{FT3D_{s}}$ and $\mathrm{KITTI_{s}}$. Training with \Gls{rs} can generalize very well when evaluated with \gls{fps}.}
	\label{Table6_Ablation_FPSvsRS}
	\begin{center}
		\resizebox{1.0\linewidth}{!}
		{
			\begin{tabular}{c|cc|cc}				
				\multirow{4}{*}{\textbf{Train$\diagdown$Test}} & \multicolumn{2}{c|}{\textbf{FPS}} & \multicolumn{2}{c}{\textbf{RS}}
				\\
				& $\mathbf{FT3D_{s}}$~\textbf{\cite{mayer2016large}} & $\mathbf{KITTI_{s}}$~\textbf{\cite{menze2015object}}
				& $\mathbf{FT3D_{s}}$~\textbf{\cite{mayer2016large}} & $\mathbf{KITTI_{s}}$~\textbf{\cite{menze2015object}}
				\\
				& \textbf{Acc3DR} & \textbf{Acc3DR} & \textbf{Acc3DR} & \textbf{Acc3DR} 
				\\	
				& [\%]            & [\%]         & [\%]            & [\%] 
				\Tstrut\Bstrut\\
				\hline	
				FPS               
				& 98.09          & 97.63     & 85.73          & 84.53    
				\Tstrut\Bstrut\\    
				\textbf{RS} 
				& \textbf{98.10}         & \textbf{97.67}            	
				& \textbf{97.51 }        & \textbf{96.53 }           
				\Tstrut\Bstrut\\
			\end{tabular}
		}
	\end{center}
\end{table}

\begin{table}[b]
	\caption{We study the effect of augmentation on $\mathrm{FT3D_{s}}$ and $\mathrm{KITTI_{s}}$. To speed up the experiments, we train and evaluate using \gls{fps}.}
	\label{Table7_Ablation_Augmentation}
	\begin{center}
		\resizebox{1.0\linewidth}{!}
		{
			\begin{tabular}{cc|cc|cc}				
				\multirow{2}{*}{Spatial} & \multirow{2}{*}{Geometry} & \multicolumn{2}{c|}{$\mathbf{FT3D_{s}}$~\textbf{\cite{mayer2016large}}} & \multicolumn{2}{c}{$\mathbf{KITTI_{s}}$~\textbf{\cite{menze2015object}}}\\
				& & \textbf{EPE3D}  & \textbf{Acc3DR} & \textbf{EPE3D} & \textbf{Acc3DR} \\	
				& & [m]             & [\%]            & [m]             & [\%] \Tstrut\Bstrut\\
				\hline	
				\xmark & \xmark      & 0.032           & 97.87         & 0.067          & 87.23            
				\Tstrut\Bstrut\\		
				\cmark  & \xmark     & 0.033           & 97.63         & 0.044          & 93.67            
				\Tstrut\Bstrut\\		
				\xmark  & \cmark     & 0.035           & 97.85         & 0.031         & 97.57            
				\Tstrut\Bstrut\\		
				\cmark  & \cmark     & 0.029           & \textbf{98.08 }        & \textbf{0.030 }    & \textbf{97.63}           
				\Tstrut\Bstrut\\
			\end{tabular}
		}
	\end{center}
\end{table}

\textbf{Impact of Augmentation:}
We examine the effect of our data augmentation (\cf \cref{Implementation}) individually.
The results of these experiments are shown in \cref{Table7_Ablation_Augmentation}.  
As mentioned before, to speed up the tests, we train with \gls{fps} and a correspondence set of $16$.

When training without any augmentation, the results are good on $\mathrm{FT3D_{s}}$, but generalize poorly to the real-world data of $\mathrm{KITTI_{s}}$.
When we randomize the initial sampling in each epoch (\ie, change the spatial locations for each epoch), the results on the $\mathrm{FT3D_{s}}$ data set drop slightly, but the accuracy and end-point error on $\mathrm{KITTI_{s}}$ are significantly improved.
We observe a similar behavior when adding only the geometric augmentation, with an even larger positive impact on $\mathrm{KITTI_{s}}$.
Both augmentation strategies together improve the overall results on both data sets and provide the best generalization from synthetic to real scenes.

\begin{figure*}[t]
	\begin{tabular}{p{0.1cm}cccc}
		& \textit{Example 1}  & \textit{Example 2} & \textit{Example 3} 
		
		\Tstrut\Bstrut\\ 
		\rotatebox[origin=c]{90}{\textit{Scene}} &
		\includegraphics[width=0.3\linewidth,valign=c]{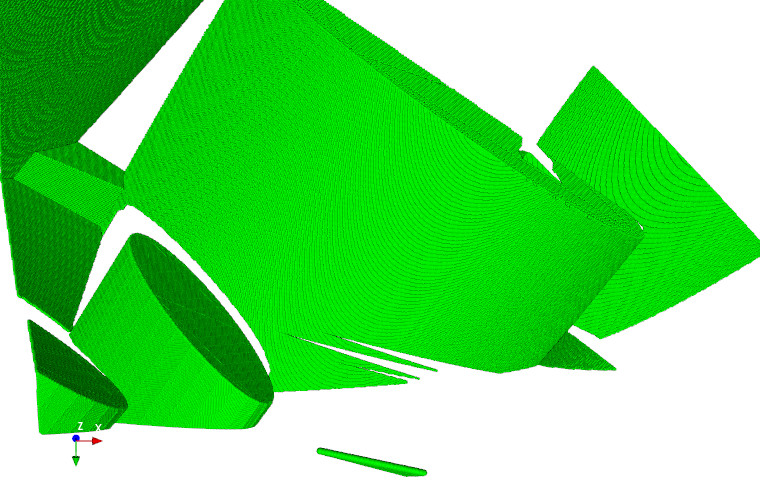}&
		\includegraphics[width=0.3\linewidth,valign=c]{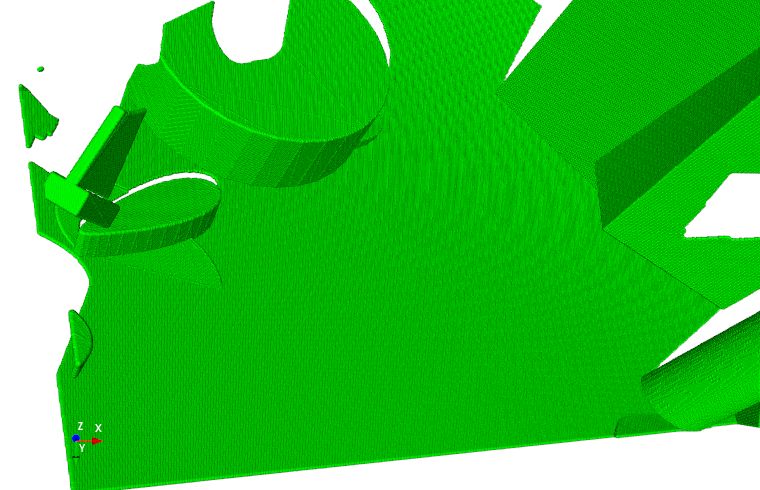}&
		\includegraphics[width=0.3\linewidth,valign=c]{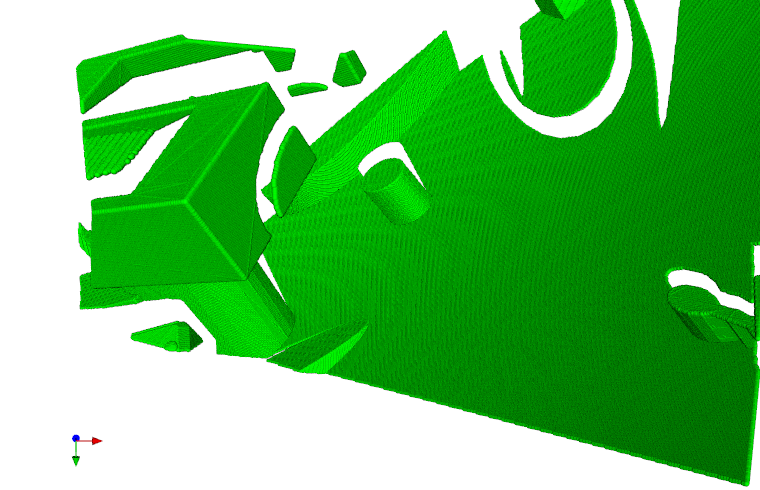}
		\Tstrut\Bstrut\\	
		
		\rotatebox[origin=c]{90}{\textit{Error (35m)}} &
		\includegraphics[width=0.3\linewidth,valign=c]{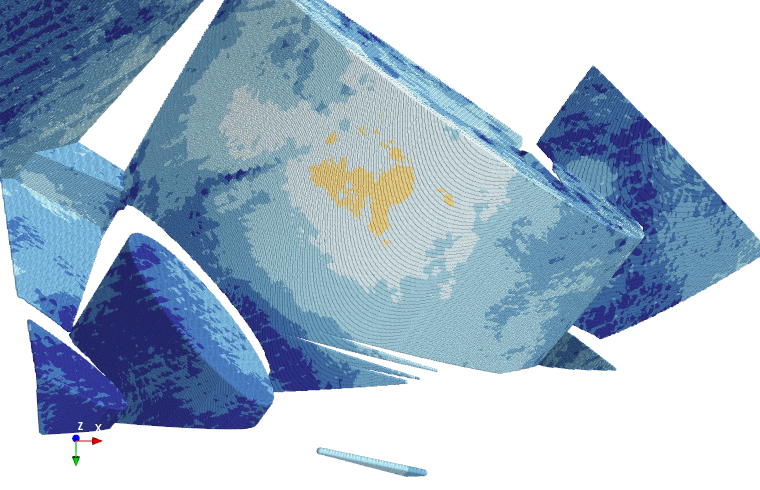}&
		\includegraphics[width=0.3\linewidth,valign=c]{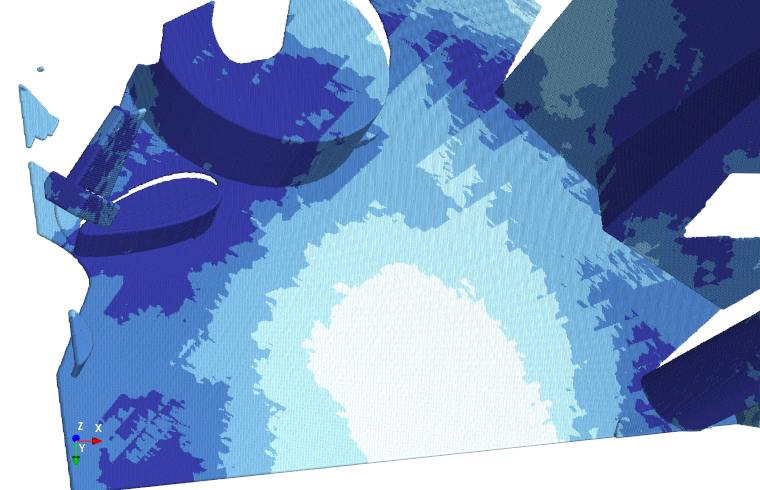}&
		\includegraphics[width=0.3\linewidth,valign=c]{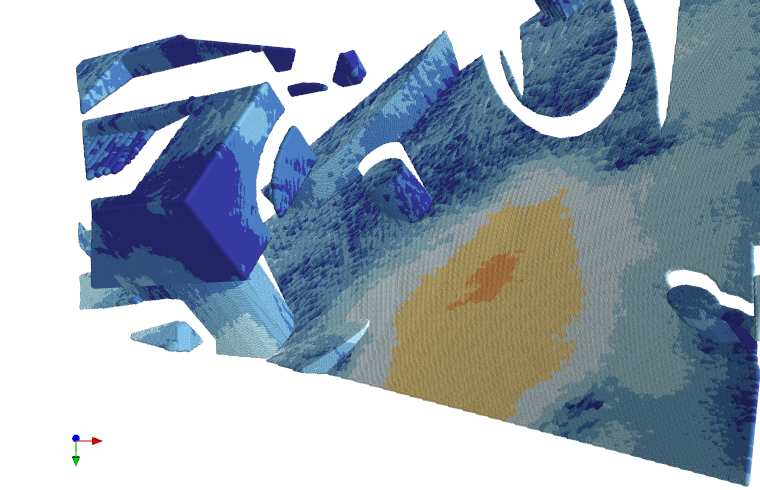}
		\Tstrut\Bstrut\\
		& \multicolumn{3}{c}{\includegraphics[width=0.95\linewidth,valign=c]{KITTI_errorcolors_3D.png}}\Tstrut\Bstrut\\
	\end{tabular}
	\caption{Three examples from the non-occluded versions of $\mathrm{FT3D_s}$ show the failure cases of our \name{} with high point densities using \gls{rs}. The scene of each example (first row) visualizes $P^t$ as green color. The error map of each scene (second row) shows the end-point error in meters according to the color map shown in the last row.}
	\label{Figure11_FailureCases}
\end{figure*}

\subsection{Limitations} \label{sec:limitations}
In terms of accuracy, there are three major limitations:
1) Any errors at the coarsest level can accumulate in the higher resolution layers, degrading the overall accuracy.
2) Our one-to-one bidirectional matching at the coarsest resolution can lead to mismatches if the scene contains repetitive patterns (\eg, along road pillars or traffic barriers). 
3) Areas of homogeneous geometry (\eg, road surfaces or grass along the road) pose a challenge to our model, especially when \gls{rs} is used (\cf \cref{Figure10_QualitativeComparison}). The error increases significantly when these untextured objects are represented or scanned by high-density points, as shown in \cref{Figure11_FailureCases}. 
In terms of efficiency, two major limitations remain: 1) To maintain accuracy at higher input densities, it is necessary to increase the resolution rates in the downsampling layers, which increases the runtime and memory requirements (\cf \cref{Figure7_AccTimeVsMethods}).
2) The~\acrshort{knn} search dominates the computational complexity as the input density increases.

\section{Conclusion}
In this paper, we propose \name{} -- an efficient and fully supervised network for multi-scale scene flow estimation in high-density point clouds. 
By using \acrfull{rs} during feature extraction, we are able to boost the runtime and memory footprint for an efficient processing of point clouds with an unmatched maximum density.
The novel \acrlong{fe} module (called \PTDP{}), resolves the prominent challenges in using \acrshort{rs} for scene flow estimation.
Compared to our preliminary work \cite{battrawy2022rms}, we reduce the operations in our network and improve the accuracy. 
We demonstrate the advantages of \gls{rs} over \gls{fps} on high-density point clouds and its ability to generalize to \gls{fps} during inference.
We provide an intensive benchmark, in which our \name{} achieves the best results in terms of accuracy, generalization, and runtime compared to the previous state-of-the-art.
We also investigate the robustness of our network to occlusions and explore its ability to operate on long-range point clouds (\ie, up to $210$ meters).

In the future, we would like to improve our model by fusing the 3D information of point clouds with textural 2D information captured by RGB cameras. 
We also plan to add ego-motion estimation to our model to avoid inaccuracies in static, homogeneous areas such as the road surface.

\vspace{4mm}

\section*{Declarations}
This version of the article has been accepted for publication, after peer review but is not the Version of Record and does not reflect post-acceptance improvements, or any corrections. The Version of Record is available online at: \url{https://doi.org/10.1007/s11263-024-02093-9}.

\bmhead{Funding}
This work was partially funded by the Federal Ministry of Education and Research Germany under the project DECODE (01IW21001), partially in the funding program Photonics Research Germany under the project FUMOS (13N16302) and partially by EU Horizon Europe Framework Program under the grant agreement 101092889 (SHARESPACE).

\bmhead{Availability of data and materials}
The data sets analyzed in the current study are available at the following links:
\begin{itemize}
\item KITTI data set: [\url{https://www.cvlibs.net/datasets/kitti/eval_scene_flow.php}]
\item FlyingThings3D and FlyingThings3D subset data sets: [\url{https://academictorrents.com/userdetails.php?id=9551}],  [\url{https://lmb.informatik.uni-freiburg.de/resources/datasets/SceneFlowDatasets.en.html}]
\end{itemize}
Other repository names of the established preprocessing scripts for the data sets with the persistent web links are available in the manuscript.

\bibliographystyle{plain}
\bibliography{bibliographyfile}
\end{document}